%% file: arXiv.tex
\newif\ifJOURNAL
\newif\ifCONF
\newif\ifarXiv
\newif\ifWP
\newif\ifFULL

\arXivtrue

\ifJOURNAL
  \documentclass[times,review,10pt]{elsarticle} 
  \usepackage{amsmath,amsthm,amsfonts,amssymb,latexsym,graphicx,etoolbox,stmaryrd,environ}
\fi
\ifCONF
  \documentclass[runningheads]{llncs}
  \usepackage{amsmath,amsfonts,latexsym,etoolbox,stmaryrd,environ}
\fi
\ifarXiv
  \documentclass[10pt]{article}
  \usepackage{amsmath,amsthm,amsfonts,amssymb,latexsym,graphicx,etoolbox,stmaryrd,environ}
\fi
\ifWP
  \documentclass[10pt]{article}
  \usepackage{amsmath,amsthm,amsfonts,amssymb,latexsym,graphicx,accents,etoolbox,stmaryrd,environ}
  \input{T2Aenc.def}
  \usepackage[utf8]{inputenc}

  \input{/Doc/Computing/Latex/kp.txt}
\fi

\usepackage[shortcuts]{extdash} 

\newif\ifTR   
\ifarXiv\TRtrue\fi
\ifWP\TRtrue\fi

\allowdisplaybreaks[4]

\newcommand{\Extra}[1]{}
\ifFULL
  \usepackage{xcolor}
  
  \newcommand{\bluebegin}{\begingroup\color{blue}}
  \newcommand{\blueend}{\endgroup}
\fi

\usepackage[pdfpagemode=UseNone,pdfstartview=FitH]{hyperref}  

\newtoggle{JOURNAL}
\newtoggle{CONF}
\newtoggle{arXiv}
\newtoggle{WP}
\newtoggle{FULL}
\newtoggle{TR}
\newtoggle{false} 
\newtoggle{true}\toggletrue{true} 

\ifJOURNAL\toggletrue{JOURNAL}\fi
\ifCONF\toggletrue{CONF}\fi
\ifarXiv\toggletrue{arXiv}\fi
\ifWP\toggletrue{WP}\fi
\ifFULL\toggletrue{FULL}\fi
\ifTR\toggletrue{TR}\fi

  \newtheorem{theorem}{Theorem}
  \newtheorem{proposition}[theorem]{Proposition}
  \newtheorem{lemma}[theorem]{Lemma}
  \newtheorem{corollary}[theorem]{Corollary}

  \theoremstyle{definition}
  
  \newtheorem{example}[theorem]{Example}
  \iftoggle{FULL}{}{}

  \theoremstyle{remark}
  \newtheorem{remark}[theorem]{Remark}

\newcommand{\Z}{\mathbb{Z}}   
\newcommand{\R}{\mathbb{R}}   

\newcommand{\dd}{\,\mathrm{d}}  
\renewcommand{\d}{\mathrm{d}}   

\renewcommand{\P}{\mathbb{P}}  
\newcommand{\E}{\mathbb{E}}    
\newcommand{\FFF}{\mathcal{F}} 
\newcommand{\GGG}{\mathcal{G}} 
\newcommand{\KKK}{\mathcal{K}} 

\newcommand{\SV}{\textrm{SV}}  
\newcommand{\VS}{\textrm{VS}}  
\newcommand{\Sh}{\textrm{Sh}}  

\DeclareMathOperator{\var}{var}   

\NewEnviron{wideformula*}{
  \iftoggle{JOURNAL}%
    {\begin{equation*}\BODY\end{equation*}}%
    {\begin{multline*}\BODY\end{multline*}}}
\NewEnviron{wideformula}{
  \iftoggle{JOURNAL}%
    {\begin{equation}\BODY\end{equation}}%
    {\begin{multline}\BODY\end{multline}}}

\iftoggle{CONF}{%
  \title{Conformal e-prediction}
  \author{Vladimir Vovk}
  \authorrunning{V.~Vovk}
  \institute{Department of Computer Science\\
    Royal Holloway, University of London\\
    \email{v.vovk@rhul.ac.uk}}
}{}%

\iftoggle{TR}{%
  \title{Conformal e-prediction}
  \author{Vladimir Vovk}
}{}%

\iftoggle{WP}{%
  
  \twodatestrue
  
}{}%

\begin{document}
\iftoggle{JOURNAL}{\begin{frontmatter}}{\maketitle}

\iftoggle{JOURNAL}{%
  \title{Conformal e-prediction}
  \author{Vladimir Vovk}
  \address{Department of Computer Science\\
    Royal Holloway, University of London\\
    Egham, Surrey TW20 0EX, UK}
  \ead{v.vovk@rhul.ac.uk}
}{}%

\begin{abstract}
  This paper discusses a counterpart of conformal prediction for e-values,
  \emph{conformal e-prediction}.
  Conformal e-prediction is conceptually simpler
  and had been developed in the 1990s
  \iftoggle{FULL}{\bluebegin
    by Gammerman, Vapnik, and Vovk
  \blueend}{}%
  as a precursor of conformal prediction.
  When conformal prediction emerged as result of replacing e-values by p-values,
  it seemed to have important advantages over conformal e-prediction
  without obvious disadvantages.
  This paper re-examines relations between conformal prediction and conformal e-prediction
  systematically from a modern perspective.
  Conformal e-prediction has advantages of its own,
  such as the ease of designing conditional conformal e-predictors
  and the guaranteed validity of cross-conformal e-predictors
  (whereas for cross-conformal predictors validity is only an empirical fact
  and can be broken with excessive randomization).
  Even where conformal prediction has clear advantages,
  conformal e-prediction can often emulate those advantages,
  more or less successfully.
  \iftoggle{FULL}{\bluebegin
    Conformal e-prediction can also serve as basis for testing;
    the resulting \emph{conformal e-testing}
    looks very different from but inherits some strengths of conformal testing.
    But this is discussed in a sister paper,
    ``Conformal e-testing''.%
  \blueend}{}%
  \iftoggle{CONF}{%
    \keywords{Conformal e-prediction \and
      cross-conformal e-prediction \and e-value.}
  }{}%
  \ifboolexpr{togl{arXiv} and not togl{JOURNAL}}{%

     The version of this paper at \url{http://alrw.net} (Working Paper 26)
     is updated most often.%
  }{}%
  \iftoggle{WP}{%


  }{}%
\end{abstract}

\iftoggle{JOURNAL}{%
  %
  \begin{keyword}
    conformal e-prediction
    \sep
    cross-conformal e-prediction
    \sep
    e-value

    \medskip

    \MSC[2020] 68T05 \sep 68Q32 \sep 62G15
  \end{keyword}}{}%

\iftoggle{JOURNAL}{\end{frontmatter}}{}

\section{Introduction}
\label{sec:introduction}

Conformal prediction is based on the notion of a p-value.
At this time p-values are widely discussed and sometimes criticized
(see, e.g., \cite{Wasserstein/etal:2019SI}),
and several alternatives to p-values have been proposed.
Perhaps the most popular alternatives are Bayes factors
and their non-Bayesian variation, e-values.
The terminology of e-values was introduced in \cite{Vovk/Wang:2021},
and the literature on e-values has been growing quickly;
see, e.g., \cite{Shafer:2021,Ramdas/etal:2023,Grunwald/etal:2024}.

\iftoggle{FULL}{\bluebegin
  In the terminology of algorithmic complexity,
  p-values are like Kolmogorov complexity,
  and e-values are like prefix complexity;
  see \cite[Appendix~A]{Vovk:arXiv2305}.
\blueend}{}%

In fact, e-values were used (under different names)
when discussing a precursor of conformal prediction in the 1990s;
in this paper we will refer to this precursor
as \emph{conformal e-prediction}.
One early description of conformal e-prediction is \cite{Gammerman/etal:1998}.
The paper \cite{Vovk/etal:1999} that first introduced conformal prediction
also discusses conformal e-prediction.
In this paper, we will occasionally refer to conformal prediction as \emph{conformal p-prediction}
in order to emphasize it being based on p-values.

Soon after the publication of \cite{Gammerman/etal:1998,Vovk/etal:1999},
conformal e-prediction seems to have disappeared.
Perhaps the main reason why it was superseded by conformal prediction
was that conformal predictions can be packaged as prediction sets
\cite[Sect.~2.2]{Vovk/etal:2022book},
and in this case their property of validity is very easy to state:
we just say that the probability of error is at most $\epsilon$
at a prespecified significance level $\epsilon$
\cite[Proposition~2.3]{Vovk/etal:2022book}.
This was clearly stated only in 2001 \cite[Theorem~1]{Nouretdinov/etal:2001ICML},
although this statement was implicit
in the standard requirement of validity for p-values stated in \cite{Vovk/etal:1999}.
The standard requirement of validity for e-values, also stated in \cite{Vovk/etal:1999},
does not admit such a simple restatement in terms of probability of error
without weakening it drastically;
see \iftoggle{JOURNAL}{\ref{app:BB}}{Appendix~\ref{app:BB}}
for further details.
(While the prediction sets derived from conformal e-prediction
can be used to define the property of validity in its strong form,
validity becomes a property of the whole family of prediction sets
for different significance levels.)

\iftoggle{FULL}{\bluebegin
  Another, perhaps less important, advantage of conformal p-prediction
  is that conformal predictors are more invariant w.r.\ to the choice of the nonconformity measure:
  they only depends on the nonconformity order rather than the nonconformity measure per se.%
\blueend}{}%

Another reason for conformal e-prediction losing its popularity
may have been the finding in 2002 \cite[Theorem~1]{Vovk:2002FOCS}
that, in the on-line mode of prediction,
smoothed conformal predictors make errors independently.
An important corollary of this stronger property of validity
is that small probabilities of errors manifest themselves,
with high probability,
as a low frequency of errors
\cite[Corollary~2.5]{Vovk/etal:2022book}.

\iftoggle{FULL}{\bluebegin
  \textbf{For the sister paper:}
  A third advantage of conformal prediction
  is that the strong property of validity found in \cite[Theorem~1]{Vovk:2002FOCS}
  allows us to test the assumption of exchangeability.
  This strong property can be stated as the independence of the conformal p-values
  output at different steps,
  and to test exchangeability,
  we can bet against the conformal p-values being independent and uniformly distributed.
  This led to the introduction in 2003 \cite{Vovk/etal:2003ICML}
  of conformal test martingales.
  (They were discussed, albeit without going beyond \cite{Vovk/etal:2003ICML},
  in the first edition of \cite{Vovk/etal:2022book}.)
  For a long time, nothing was known about the efficiency of conformal test martingales,
  and first results about their efficiency appeared in 2019
  (see \cite{Vovk:2021-full});
  this is a major topic of \cite[Part~III]{Vovk/etal:2022book}.%
\blueend}{}%

The last advantage of conformal prediction that we discuss in this section
was found only in 2017 \cite{Vovk/etal:2019ML},
and so it did not contribute to the eclipse of conformal e-prediction
in the early 2000s.
It was the discovery of conformal predictive distributions,
motivated by \cite{Shen/etal:2018}:
in the case of regression,
smoothed conformal prediction may produce ``conformal predictive distributions'',
which are automatically well-calibrated.

In this paper we will look systematically at these advantages of conformal prediction
except for the last one,
which will only be briefly discussed in the concluding section.
On one hand, the favourable properties of conformal prediction
are often partially satisfied by conformal e-prediction.
And on the other hand,
conformal e-prediction has several advantages of its own.

For simplicity,
in \iftoggle{FULL}{\bluebegin most of \blueend}{}this paper
\iftoggle{FULL}{\bluebegin(Sect.~\ref{sec:OCM} is the only exception) \blueend}{}%
we consider IID (or at least exchangeable) data
and concentrate on the problem of pattern recognition
(also known as classification).
\iftoggle{FULL}{\bluebegin
  We mention regression only occasionally.
\blueend}{}%
We start in Sect.~\ref{sec:CeP} from the definition of conformal e-prediction
and continue in Sect.~\ref{sec:validity} with discussing its validity.
The simplest property of validity
(Proposition~\ref{prop:space} in Sect.~\ref{sec:validity})
consists in conformal e-predictors producing at each step
a valid e-value for the true label:
namely, its expectation is at most 1 at each step;
the general notion of e-value is introduced right after Proposition~\ref{prop:space}.
Since the expectation is the average over the sample space,
we can say that conformal e-predictors are valid in the space domain, or space-wise.
A complementary notion of efficiency is efficiency in the time domain,
which we consider next.

Proposition~\ref{prop:space} does not say anything
about the relation between the e-values for the true labels produced at different steps.
Is it possible that for some streams of data
the average of the e-values produced at different steps tends to 2
while for others it tends to 0.5 with non-zero probability?
In Sect.~\ref{sec:validity} we show that this is impossible in the online prediction protocol,
stating both the strong law of large numbers and the law of the iterated logarithm
for the e-values produced for the true labels at different steps.
This does not fully replace the strong property of independence of errors
for conformal prediction,
but it can be regarded as a partial replacement.
We can see that the e-values produced at different steps
are not misleading in the time domain;
not only is their expectation at most 1 at each step,
their average is at most 1 time-wise in the long run.

The properties of validity established in Sect.~\ref{sec:validity}
are marginal, in the sense that the expectations, or time averages, in them
are not conditional on any properties of the observations.
For example, if our predictions are for people,
in principle we can get very different averages for men and women.
In the case of conformal prediction,
a simple way of achieving conditional validity is using Mondrian conformal prediction
\cite[Sect.~4.6]{Vovk/etal:2022book}.
\iftoggle{FULL}{\bluebegin
  Interestingly,
  Mondrian conformal prediction was first published
  only in our 2005 book \cite{Vovk/etal:2005book}.
  Before that, it was only posted as Working Paper 4 in the OCP Old Series.
\blueend}{}%
Mondrian conformal prediction requires a hard partition of observations,
such as the partition of people into men and women.
Interestingly, conformal e-prediction is much more flexible
when trying to achieve conditional validity.
Its conditional version does not have to be based on a partition (Sect.~\ref{sec:conditional});
e.g., we may require separate validity for men, women, and Europeans.
It has been shown recently that this type of conditionality
can also be achieved for conformal prediction
(see, e.g., \cite{Gibbs/etal:arXiv2305}),
but it is much easier to achieve and appears to be more natural
in the case of conformal e-prediction.
This can be regarded as an advantage of conformal e-prediction.

What we discuss in Sections~\ref{sec:CeP}--\ref{sec:conditional} is ``full'' conformal e-prediction,
and it is computationally inefficient
when built on top of many standard prediction algorithms
(such as neural networks).
Section~\ref{sec:split} introduces split conformal e-prediction,
which is a simple way to make conformal e-prediction computationally efficient.
Similarly to split-conformal prediction
(introduced in \cite{Papadopoulos/etal:2002ICMLA,Papadopoulos/etal:2002ECML}),
split conformal e-prediction
can lose in predictive efficiency as compared with full conformal e-prediction.

To prevent loss in computational efficiency without sacrificing predictive efficiency,
cross-conformal predictors were introduced in \cite{Vovk:2015-cross}.
\iftoggle{FULL}{\bluebegin
  It is interesting that this paper was never published in conference proceedings
  or as an arXiv report.
\blueend}{}%
Cross-conformal predictors are not provably valid \cite[Appendix]{Vovk:2015-cross},
and this sometimes even shows in experimental results \cite{Linusson/etal:2017-local}.
The limits of violations of validity are given by R\"uschendorf's result
(see, e.g., \cite[Proposition~2]{Vovk/Wang:2020}):
when merging p-values coming from different folds by taking arithmetic mean
(this is essentially what cross-conformal predictors do),
the resulting arithmetic mean has to be multiplied by 2
in order to guarantee validity.
In the more recent method of jackknife+, introduced in \cite{Barber/etal:2021}
and closely related to cross-conformal prediction,
there is a similar factor of 2 \cite[Theorem~1]{Barber/etal:2021},
which cannot be removed in general \cite[Theorem~2]{Barber/etal:2021}.

In Sect.~\ref{sec:cross},
we introduce a version of cross-conformal prediction based on e-values,
which we call \emph{cross-conformal e-prediction}.
The situation with cross-conformal e\-/prediction is very different
from cross-conformal prediction,
as the arithmetic mean of e-values is always an e-value.
This is an obvious fact, and it is shown in \cite[Sect.~3]{Vovk/Wang:2021}
that arithmetic mean is the only useful merging rule.
Therefore, cross-conformal e\-/prediction is always valid.
This is a second advantage of conformal e-prediction.

The emphasis of Sections~\ref{sec:validity}--\ref{sec:cross}
is on the validity of conformal e-prediction and its computational efficiency,
while Sect.~\ref{sec:efficiency} moves on to its predictive efficiency.
What are suitable criteria of predictive efficiency?
We propose two such criteria in the case of pattern recognition,
the ``observed log criterion'' and the ``prior log criterion''.
\iftoggle{FULL}{\bluebegin
  In typical regression problems, both criteria reduce to one.%
\blueend}{}%

\iftoggle{FULL}{\bluebegin
  While most of Sections~\ref{sec:CeP}--\ref{sec:no-CPD}
  rely on the exchangeability assumption,
  in Sect.~\ref{sec:OCM} we embed the exchangeability model
  into the wide class of online compression models,
  introduced in the form used here in 2003 \cite{Vovk:2006}
  but known in statistics in different forms for a long time.
  The emphasis of this section will be on the classical Gaussian model,
  and we will explore experimentally natural variations
  of Student's t-test.%
\blueend}{}%

Section~\ref{sec:conclusion} concludes and lists some advantages and disadvantages
of conformal e-prediction as compared with conformal prediction.

\iftoggle{FULL}{\bluebegin
  The foundations of probability used in this paper
  (analogously to \cite{Vovk/etal:2022book}):
  \begin{itemize}
  \item
    Measure-theoretic, but I am using canonical rather than abstract probability spaces,
    such as $\mathbf{Z}^N$;
  \item
    I never use reverse martingales;
    instead, I use martingales with negative indices
    (as already Doob \cite[Theorems~4.2 and~4.3]{Doob:1953} did).
  \end{itemize}
\blueend}{}%

\section{Conformal e-predictors}
\label{sec:CeP}

Suppose we are given a training set $z_1,\dots,z_n$ consisting of labelled objects $z_i=(x_i,y_i)$
and our goal is to predict the label of a new object $x$.
In this paper we consider predictors of the following type:
for each potential label $y$ for $x$ we would like to have a number $f(z_1,\dots,z_n,x,y)$
reflecting the plausibility of $y$ being the true label of $x$.
An example is conformal transducers \cite[Sect.~2.7]{Vovk/etal:2022book},
which, in the terminology of this paper, may be called \emph{conformal p-predictors}.
The output
\[
  y \mapsto f(z_1,\dots,z_n,x,y)
\]
of a conformal p-predictor is the full conformal prediction for the label of $x$;
e.g., it determines the prediction set at each significance level.
We will sometimes write $f(z_1,\dots,z_n,z)$, where $z:=(x,y)$,
instead of $f(z_1,\dots,z_n,x,y)$.

We will use the notation $\mathbf{X}$ for the object space and $\mathbf{Y}$ for the label space
(both assumed non-empty).
These are measurable spaces from which the objects and labels, respectively, are drawn.
Full observations $z=(x,y)$ are drawn from the observation space $\mathbf{Z}:=\mathbf{X}\times\mathbf{Y}$.
For any non-empty set $X$,
$X^+$ will be the set $\cup_{n=1}^{\infty}X^n$ of all non-empty finite sequences of elements of $X$.

A \emph{nonconformity e-measure} is a measurable function $A:\mathbf{Z}^+\to[0,\infty)^+$
that maps any finite sequence $(z_1,\dots,z_m)$, $m\in\{1,2,\dots\}$,
to a finite sequence $(\alpha_1,\dots,\alpha_m)$ of the same length
consisting of nonnegative numbers (\emph{nonconformity scores})
with average at most 1,
\begin{equation}\label{eq:1}
  \frac1m
  \sum_{i=1}^m
  \alpha_i
  \le
  1,
\end{equation}
that satisfies the following property of equivariance:
for any $m\in\{2,3,\dots\}$, any permutation $\pi$ of $\{1,\dots,m\}$,
any $(z_1,\dots,z_m)\in\mathbf{Z}^m$,
and any $(\alpha_1,\dots,\alpha_m)\in[0,\infty)^m$,
\[
  (\alpha_1,\dots,\alpha_m) = A(z_1,\dots,z_m)
  \Longrightarrow
  (\alpha_{\pi(1)},\dots,\alpha_{\pi(m)}) = A(z_{\pi(1)},\dots,z_{\pi(m)}).
\]
Very roughly, this property means that each nonconformity score $\alpha_i$
(supposed to measure the strangeness of $z_i$ as compared
with the other observations in the sequence)
does not depend on the position of $z_i$, or of the other observations, in the sequence.
The \emph{conformal e-predictor} $f$ corresponding to such $A$ is defined by
\begin{equation}\label{eq:CeP-1}
  f(z_1,\dots,z_n,x,y)
  :=
  \alpha_{n+1},
  \text{\enspace where\enspace}
  (\alpha_1,\dots,\alpha_n,\alpha_{n+1}) := A(z_1,\dots,z_n,(x,y)),
\end{equation}
so that $f:\mathbf{Z}^+\to[0,\infty)$.
A \emph{conformal e-predictor} is a function that can be obtained
from a nonconformity e-measure in this way.

When given a training set $z_1,\dots,z_n$ and a test object $x$,
the full prediction for $x$ according to a conformal e-predictor $f$
is the family of \emph{conformal e-values}
\begin{equation}\label{eq:e-family}
  \left(
    f(z_1,\dots,z_n,x,y)
    \mid
    y\in\mathbf{Y}
  \right).
\end{equation}
We can regard the family \eqref{eq:e-family} of e-values
for each potential label $y$ as a soft set predictor.
By thresholding $f$ at some level,
we can get a (hard) set predictor (as in \eqref{eq:BB-general} below);
we include in the prediction set for the label of $x$
the potential labels $y$ for which the conformal e-value is less than a chosen level.
But there is no need to choose the level in advance,
and we interpret the conformal e-value $f(z_1,\dots,z_n,x,y)$ of $y$
as the degree to which $y$ is excluded from the soft prediction set.
We want our conformal prediction to be valid and efficient,
where validity means that we do not want the true label to be excluded
(i.e., to have a large e-value)
while efficiency means that we want to exclude all other labels.

The full prediction \eqref{eq:e-family} for the label of $x$ can be summarized as, e.g.,
the \emph{point prediction}
\[
  \hat y\in\arg\min_y f(z_1,\dots,z_n,x,y)
\]
(assuming the $\min$ is attained at a single label),
the \emph{e-confidence}
\[
  \min_{y\ne\hat y}f(z_1,\dots,z_n,x,y),
\]
and the \emph{e-credibility} $f(z_1,\dots,z_n,x,\hat y)$.
(See \cite[Sect.~3.5.1]{Vovk/etal:2022book} for their p-counterparts.)
We can make a confident point prediction
when the e-confidence is large while the e-credibility is not.

Let us say that a nonconformity e-measure
and the corresponding conformal e-predictor are \emph{admissible}
if we always have ``$=$'' in place of ``$\le$'' in the definition \eqref{eq:1}.
If a conformal e-predictor is not admissible,
we can make its predictions more confident without sacrificing their validity.
Therefore, we will usually concentrate on admissible nonconformity e-measures
and admissible conformal e-predictors.

A \emph{nonnegative nonconformity measure} $A:\mathbf{Z}^+\to[0,\infty)^+$
is defined as a nonconformity e-measure
except that the condition \eqref{eq:1} is omitted.
Given a nonnegative nonconformity measure $A$,
we can always define the corresponding admissible nonconformity e-measure $A'$
by normalizing $A$:
\begin{equation}\label{eq:normalizing}
  A'(z_1,\dots,z_m)
  :=
  \frac{m}{\sum_{i=1}^m\alpha_i}
  (\alpha_1,\dots,\alpha_m),
\end{equation}
where $(\alpha_1,\dots,\alpha_m) := A(z_1,\dots,z_m)$;
if $A(z_1,\dots,z_m)=(0,\dots,0)$,
we set $A'(z_1,\dots,z_m):=(1,\dots,1)$ in order to ensure that $A'$ is admissible.
We will say that the corresponding conformal e-predictor
\emph{is based on $A$}.

A further generalization of nonnegative nonconformity measures,
\emph{nonconformity measures}, is obtained by dropping the condition of nonnegativity;
these are functions $A:\mathbf{Z}^+\to\R^+$.
\iftoggle{FULL}{\bluebegin
  Sometimes it is convenient to drop ``nonnegative''
  for specific nonconformity measures,
  such as the identical nonconformity measures below.
\blueend}{}%
They are used, explicitly or implicitly, in conformal p-prediction.
See, e.g., \cite[Sect.~1.3]{Bala/etal:2014}
(and Remark~\ref{rem:R1-4} below).

The conformal e-predictor proposed in \cite{Gammerman/etal:1998}
for binary pattern recognition problems (with $\mathbf{Y}=\{-1,1\}$)
is based on support vector machines (SVM);
let us fix all relevant parameters, such as the kernel.
It is defined as
\begin{equation}\label{eq:SVM}
  f(z_1,\dots,z_{n+1})
  :=
  \begin{cases}
    (n+1)/\left|\SV\right| & \text{if $n+1\in\SV$}\\
    0 & \text{otherwise},
  \end{cases}
\end{equation}
where $\SV$ is the set of indices of support vectors:
$i\in\SV$ if and only if $z_i$, $i\in\{1,\dots,n+1\}$, is a support vector
for the SVM constructed from $z_1,\dots,z_{n+1}$ as training set.
It is based on the indicator function of being a support vector;
indeed, the right-hand side of \eqref{eq:normalizing} then becomes
\[
  \left(
    \frac{(n+1)1_{\{i\in\SV\}}}{\sum_{i=1}^{n+1}1_{\{i\in\SV\}}}
  \right)_{i=1}^{n+1},
\]
which agrees with \eqref{eq:SVM}.
When given a training set $z_1,\dots,z_n$ and a new object $x$,
this conformal e-predictor goes through all potential labels $y$ for $x$
and for each constructs an SVM and outputs $f(z_1,\dots,z_n,x,y)$.
It makes it computationally inefficient.

\begin{remark}\label{rem:R1-4}
  In conformal prediction,
  several natural ways of defining nonconformity measures
  have been discussed and widely used in literature.
  The definition at the beginning of this section
  parallels the definition of nonconformity measures in \cite[Sect.~1.3]{Bala/etal:2014};
  under these definitions,
  nonconformity measures map sequences of observations to sequences of nonconformity scores.
  Under other definitions,
  nonconformity measures may map, e.g., a bag of observations and another observation $z$
  to one nonconformity score (that for $z$),
  and there are several varieties of such definitions;
  see, e.g., \cite[Sect.~2.9.3]{Vovk/etal:2022book}.
  Each of these varieties could have been adapted to conformal e-prediction.
\end{remark}

\section{Validity of conformal e-predictors}
\label{sec:validity}

The following obvious proposition asserts the validity of conformal e-predictors.
Let us write $Z_1,Z_2,\dots$ for the random elements
whose realizations are the observed data $z_1,z_2,\dots$;
more generally, $(X,Y)$ or $Z$ are random elements
with values in the observation space $\mathbf{Z}$.
As usual, a finite sequence of random elements is \emph{exchangeable}
if its joint distribution does not change if it is permuted
(and an infinite sequence is exchangeable if its joint distribution
does not change if its first $n$ elements are permuted, for any $n$).
The difference between exchangeability and being IID (independent and identically distributed)
can be substantial for finite data sequences
(but for infinite data sequences the difference between the two assumptions disappears,
provided $\mathbf{Z}$ is a standard Borel space,
according to de Finetti's theorem \cite[Sect.~A.5.1]{Vovk/etal:2022book}).

\begin{proposition}\label{prop:space}
  For any conformal e-predictor $f$ and any $n$,
  if $Z_1,\dots,Z_n,(X,Y)$ are IID (or exchangeable),
  \begin{equation}\label{eq:space}
    \E f(Z_1,\dots,Z_n,X,Y) \le 1
  \end{equation}
  (with ``$=$'' in place of ``$\le$'' if $f$ is admissible).
\end{proposition}

\begin{proof}
  It follows from the definition of conformal e-predictors that
  \[
    \E
    \left(
      f(Z_1,\dots,Z_n,X,Y)
      \mid
      \lbag Z_1,\dots,Z_n,(X,Y)\rbag
    \right)
    \le
    1,
  \]
  and it remains to average over the multisets $\lbag Z_1,\dots,Z_n,(X,Y)\rbag$.
\end{proof}

The property of validity given in Proposition~\ref{prop:space}
says that conformal e-predictors output valid e-values for the true labels.
Formally, an \emph{e-variable} is a random variable $E$ satisfying $\E(E)\le1$
under the data-generating distribution
(and the values it takes are \emph{e-values}).
We do not expect the e-values for the true labels to be large
because, by Markov's inequality, $\P(E\ge C)\le1/C$ for any constant $C>1$.

Proposition~\ref{prop:space} involves a ``space average'',
i.e., an average over the sample space.
The analogous property of validity for conformal prediction
can be stated in terms of error probabilities for set predictions
(see, e.g., \cite[Proposition 2.1]{Vovk/etal:2022book}),
which is particularly intuitive.
This is the first advantage of conformal prediction pointed out in Sect.~\ref{sec:introduction}.
It disappears when we move to conformal e-prediction:
validity has to be defined in a more complicated way
in order to avoid making this property much weaker
(however, it has been argued \cite{Shafer:2021} that e-values are more intuitive than p-values).

Another advantage of conformal prediction is that,
in the online mode of prediction (to be defined shortly),
conformal predictors (in their smoothed version) make errors at different steps independently
(see, e.g., \cite[Theorem~11.1]{Vovk/etal:2022book}).
Without independence, it is possible,
even when the probability of error at each step is $\epsilon$,
for the long-term relative frequency of errors over consecutive steps
to be either 0 or 1 (1 with probability $\epsilon$).
The independence of errors forces the long-term frequency of errors
to be $\epsilon$ almost surely
(which is also asserted in \cite[Proposition 2.1]{Vovk/etal:2022book}).
We can say that independence implies ergodicity,
i.e., the almost sure coincidence of time and space averages.

The independence of conformal p-values in the online mode of prediction
is the strongest property of validity in conformal prediction
as presented in \cite{Vovk/etal:2022book}.
In conformal e-prediction, independence is lost,
but ergodicity still holds, at least for bounded conformal e-predictors.

To give an example demonstrating the loss of independence
in conformal e-prediction,
we first define the online prediction protocol.
In the online protocol for conformal e-prediction,
we observe an object $x_1$,
apply the conformal e\-/predictor to compute the conformal e-values $f(x_1,y)$
for all possible labels $y\in\mathbf{Y}$,
observe the true label $y_1$,
record the e-value $e_1:=f(x_1,y_1)$ for it,
observe another object $x_2$,
apply the conformal e-predictor to compute the e-values $f(x_1,y_1,x_2,y)$
for all possible labels $y\in\mathbf{Y}$,
observe the true label $y_2$,
record the e-value $e_2:=f(x_1,y_1,x_2,y_2)$ for it,
etc.
In the case of conformal prediction,
we can instead record conformal p-values $p_n$
(or record whether a mistake is made at a given significance level).
While the conformal p-values are independent (for smoothed conformal predictors),
the following example shows that the conformal e-values are not independent
already in a toy binary situation
(perhaps this is the simplest non-trivial example of conformal e-prediction,
albeit it is not particularly interesting per se).
We write $E_n$ for the e-value $e_n$ regarded as a random variable
(an e-variable).

\begin{example}
  Let the random observations $Z_1,Z_2,\dots$ correspond to tossing a fair coin:
  $Z_n\in\{0,1\}$ and $Z_n=1$ with probability $1/2$ independently.
  Let $A$ be the identity nonconformity measure,
  \[
    A(z_1,\dots,z_m)
    :=
    (z_1,\dots,z_m),
  \]
  and consider the conformal e-predictor based on $A$
  (via \eqref{eq:normalizing}).
  Then the e\-/variables that it outputs satisfy
  \begin{equation}\label{eq:e}
    E_{n+1}
    =
    \begin{cases}
      0 & \text{with probability $1/2$}\\
      \frac{n+1}{k+1} & \text{with probability $1/2$}
    \end{cases}
  \end{equation}
  given $Z_1,\dots,Z_n$,
  where $k:=\sum_{i=1}^n Z_i$.
  This assumes $(Z_1,\dots,Z_n)\ne0$;
  if $(Z_1,\dots,Z_n)=(0,\dots,0)$, the 0 in \eqref{eq:e} should be replaced by 1.
  Since $E_1,\dots,E_n$ uniquely determine $Z_1,\dots,Z_n$
  unless $(E_1,\dots,E_n)=(1,\dots,1)$,
  we still have \eqref{eq:e} given $E_1,\dots,E_n$
  such that $(E_1,\dots,E_n)\ne(1,\dots,1)$.
  Therefore,
  \begin{equation*} 
    \E(E_{n+1}\mid E_1,\dots,E_n)
    =
    \frac{n+1}{2(k+1)}
  \end{equation*}
  provided $(E_1,\dots,E_n)\ne(1,\dots,1)$,
  and the expression on the right-hand side
  may well be different from 1
  (albeit will typically be close to 1).
  This shows that the e-values are not independent:
  their conditional expectations may be different
  from their marginal expectations.%
  \iftoggle{FULL}{\bluebegin

    Let us check that the expectation of the expectations
    $\E(e_{n+1}\mid e_1,\dots,e_n)$
    is correct.
    We have:
    \begin{multline*}
      \sum_{k=0}^n
      \frac{n+1}{2(k+1)}
      \binom{n}{k}
      2^{-n}
      =
      \sum_{k=0}^n
      \frac{n+1}{2(k+1)}
      \frac{n!}{k!(n-k)!}
      2^{-n}\\
      =
      \sum_{k=0}^n
      \frac{(n+1)!}{(k+1)!(n-k)!}
      2^{-n-1}
      =
      \sum_{k=0}^{n+1}
      \frac{(n+1)!}{k!(n-k)!}
      2^{-n-1}
      -
      \frac{(n+1)!}{0!(n+1)!}
      2^{-n-1}\\
      =
      1
      -
      2^{-n-1},
    \end{multline*}
    and the last expression is exactly the probability of not obtaining
    \[
      (z_1,\dots,z_{n+1})=(0,\dots,0),
    \]
    which leads to $e_{n+1}=0$.%
  \blueend}{}%
\end{example}

Nevertheless, various properties of ergodicity still hold.
For example,
the following proposition asserts a simple property of ergodicity
for the conformal e-values $e_n:=f(x_1,y_1,\dots,x_n,y_n)$,
namely their asymptotic online validity.

\begin{proposition}\label{prop:time}
  Suppose the observations $(X_n,Y_n)$, $n=1,2,\dots$, are IID
  and e-variables $E_n$ are produced by a bounded conformal e-predictor $f$.
  Then, in the online prediction protocol,
  \begin{equation}\label{eq:goal}
    \limsup_{N\to\infty}
    \frac1N
    \sum_{n=1}^N
    E_n
    \le
    1
    \quad\text{a.s.}
  \end{equation}
  We can replace the ``$\le$'' by ``$=$'' if $f$ is admissible.
\end{proposition}

\noindent
Proposition~\ref{prop:time} shows that the long-term time average of the e-values for the true labels
is bounded above by 1 almost surely.
In this sense they are time-wise e-values.

\begin{proof}[Proof of Proposition~\ref{prop:time}]
  We follow the proof of \cite[Lemma~14]{Vovk:2004}
  (given as Lemma~3.15 in the first edition of \cite{Vovk/etal:2022book})
  and use the terminology of \cite[Chap.~7]{Shiryaev:2019}.

  Let $\FFF_n$ be the $\sigma$-algebra generated
  by the multiset $\lbag Z_1,\dots,Z_{n-1}\rbag$
  and the observations $Z_n,Z_{n+1},\dots$.
  (These $\sigma$-algebras form the \emph{exchangeable filtration}
  \cite[Sect.~5.6]{Ramdas/etal:2023}.)
  For each time horizon $N\in\{2,3,\dots\}$,
  the stochastic sequence $(E_n-1,\FFF_n)$, $n=N,\dots,1$,
  is a bounded supermartingale difference.
  By Hoeffding's inequality (see, e.g., \cite[Sect.~A.6.3]{Vovk/etal:2022book}),
  for any $N\in\{2,3,\dots\}$,
  \begin{equation}\label{eq:bound}
    \P
    \left\{
      \frac1N
      \sum_{n=1}^N
      (E_n-1)
      \ge
      \epsilon
    \right\}
    \le
    e^{-2\epsilon^2N/C^2},
  \end{equation}
  where $C$ is an upper bound on the conformal e-predictor $f$.
  By the Borel--Cantelli lemma \cite[Sect.~2.10, part (a) of the lemma]{Shiryaev:2016},
  the internal inequality in~\eqref{eq:bound}
  holds only for finitely many $N$ for a fixed $\epsilon>0$.
  This implies~(\ref{eq:goal}).

  If $f$ is admissible,
  we can also apply the same argument to $1-E_n$ in place of $E_n-1$.
\end{proof}

The equation~\eqref{eq:bound} in the proof of Proposition~\ref{prop:time}
can also be interpreted directly,
and its advantage is that it is non-asymptotic.
It implies that, for any time horizon $N\ge2$,
\begin{equation}\label{eq:finite}
  \P
  \left\{
    \frac1N
    \sum_{n=1}^N
    E_n
    \ge
    1+\epsilon
  \right\}
  \le
  e^{-2\epsilon^2N/C^2},
\end{equation}
which is another expression for the $e_n$ being time-wise e-values:
their average is approximately bounded above by 1 with high probability
(for small $\epsilon$ and large~$N$).

Even if we do not assume that the conformal e-predictor is bounded,
we can still claim, e.g., that
\begin{equation}\label{eq:truncated}
  \P
  \left\{
    \frac1N
    \sum_{n=1}^N
    (E_n\wedge N^{1/3})
    \ge
    1+\epsilon
  \right\}
  \le
  e^{-2\epsilon^2N^{1/3}}.
\end{equation}
The last inequality says that $E_n\wedge N^{1/3}$ are approximate time-wise e-values
(assuming that $\epsilon$ is small and $N\gg\epsilon^{-6}$),
and so $E_n$ are approximate time-wise e-values if we regard an e-value of $N^{1/3}$
as large enough for the difference between $N^{1/3}$ and larger e-values
to be considered unimportant
(such as an e-value of 100 on Jeffreys's scale \cite[Appendix~B]{Jeffreys:1961}).
(To derive~\eqref{eq:truncated} from~\eqref{eq:finite},
just set $C:=N^{1/3}$.)

\iftoggle{FULL}{\bluebegin
  For a fixed time horizon $N$, we can just apply results for forward martingales to $E_N,\dots,E_1$.

  See Sect.~10.16 of \cite{Gut:2013}.%
\blueend}{}%

\begin{figure}[htbp]
  \begin{center}
    \iftoggle{JOURNAL}{%
      \includegraphics[width=0.6\textwidth]{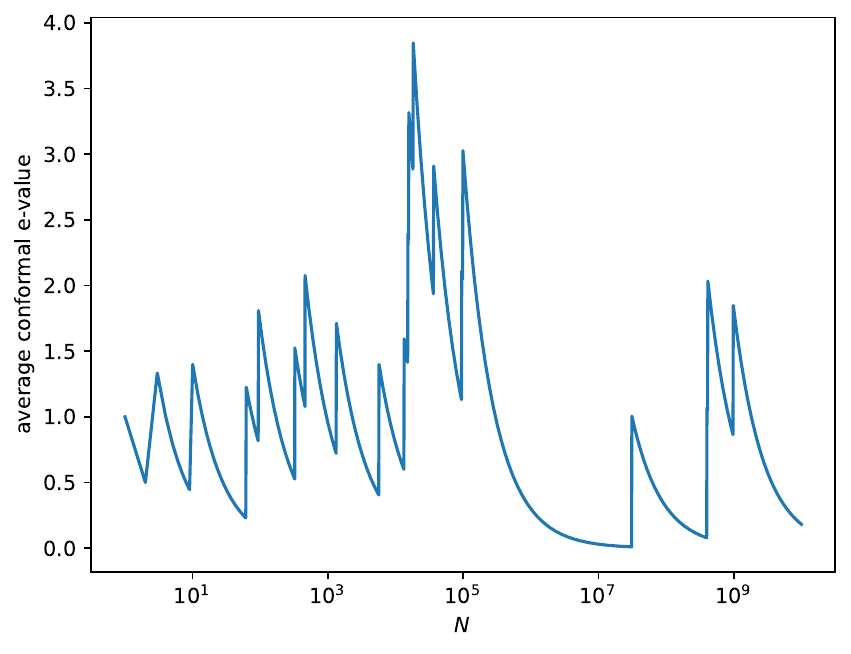}}{%
      \includegraphics[width=0.8\textwidth]{counterexample.pdf}}
  \end{center} 
  \caption{Illustration for Remark~\ref{rem:bounded}}
  \label{fig:illustration}
\end{figure}

\begin{remark}\label{rem:bounded}
  Let us check that we cannot simply drop the requirement
  that $E$ be bounded in Proposition~\ref{prop:time}.
  Suppose that, for each $n$,
  \[
    E_n
    :=
    \begin{cases}
      n & \text{with probability $\frac{1}{n}$}\\
      0 & \text{with probability $\frac{n-1}{n}$}
    \end{cases}
  \]
  independently of $E_{n+1},E_{n+2},\dots$.
  (For example, $\lvert\mathbf{X}\rvert=1$, $\mathbf{Y}=\R$,
  each observation is generated from the same continuous probability measure independently,
  and the nonconformity e-measure is $n$ at the largest $y_n$, assuming it is unique, and $0$ elsewhere.)
  Then the distribution of $E_1,\dots,E_N$ is the product distribution
  \begin{equation}\label{eq:product}
    \prod_{n=1}^N
    \left(
      \frac{n-1}{n} \delta_0
      +
      \frac{1}{n} \delta_n
    \right)
  \end{equation}
  (where $\delta_a$ is the probability measure concentrated on $\{a\}$),
  since $E_N,\dots,E_1$ are generated independently.
  This gives us a product distribution for $E_1,E_2,\dots$,
  namely \eqref{eq:product} with $N$ replaced by $\infty$.
  For any $b\in\{1,2,\dots\}$,
  by Le Cam's version of Poisson's theorem \cite[Sect.~3.12]{Shiryaev:2016},
  the probability of the event that $b$ of the $E_n$ for $n\in[2^k,2^{k+1})$
  will satisfy $E_n=n$ tends to $e^{-\lambda}\lambda^{b}/b!$ as $k\to\infty$,
  where $\lambda:=\ln2$;
  therefore, this event will happen infinitely often (for any $b$).
  \iftoggle{FULL}{\bluebegin
    Let us check the expression for $\lambda$.
    The cumulative probability of the events $E_n=n$ over all $n\in[2^k,2^{k+1})$ is
    \[
      \sum_{n=2^k}^{2^{k+1}-1}
      \frac1n
      \sim
      \ln(2^{k+1}) - \ln(2^{k})
      =
      \ln 2.
    \]
  \blueend}{}%
  Therefore, \eqref{eq:goal} will be violated almost surely;
  namely, the limsup in it will be $\infty$ almost surely.
  Figure~\ref{fig:illustration} illustrates this
  by plotting the average conformal e-value $\frac1N\sum_{n=1}^N e_n$ vs $N$
  for a wide range of $N$, $N\in\{1,\dots,10^{10}\}$ (with $N$ given on a logarithmic scale).
\end{remark}

In \iftoggle{JOURNAL}{\ref{app:LIL}}{Appendix~\ref{app:LIL}}
we will check that Proposition~\ref{prop:time}
can be strengthened to the following statement of the iterated-logarithm type.

\begin{proposition}\label{prop:time-LIL}
  Suppose the observations $(X_n,Y_n)$, $n=1,2,\dots$, are IID.
  Then, in the online prediction protocol,
  \begin{equation*} 
    \limsup_{N\to\infty}
    \sqrt{\frac{N}{\ln\ln N}}
    \left(
      \frac1N
      \sum_{n=1}^N
      E_n
      -
      1
    \right)
    \le
    2^{-1/2} C
    \quad\text{a.s.},
  \end{equation*}
  where $C$ is an upper bound on the conformal e-predictor
  producing $E_1,E_2,\dots$.
\end{proposition}


In conclusion of this section,
let me state what can be considered to be the main property of validity
for conformal e-prediction
(it implies the other properties listed here,
although by itself it is not particularly intuitive):
$\E(E_n\mid\FFF_{n+1})\le1$,
where $\FFF$ is the exchangeable filtration
and $E_n$ is a conformal e\-/variable.
It was introduced in the proof of Proposition~\ref{prop:time}
but is worth stating separately.

\section{Conditional conformal e-predictors}
\label{sec:conditional}

Conformal prediction and conformal e\-/prediction
satisfy the marginal property of validity that we have discussed so far.
In this section we will discuss stronger properties of validity,
including what we called \emph{object-conditional} and \emph{label-conditional} validity
in \cite[Sect.~4.6]{Vovk/etal:2022book}.
Suppose, e.g., that we have an algorithm for diagnosing Covid.
In object-conditional validity,
we divide the objects into separate categories
(such as men and women)
and require validity inside each of the categories, not just on average.
In label-conditional validity,
we require validity for each possible label, in this case separate validity
for people who have Covid and those who do not.
More generally, we can talk about \emph{observation-conditional validity},
where the categories for which we require separate validity
are defined in terms of both objects and labels.

The most standard approach to conditional conformal prediction
is Mondrian conformal prediction
(see, e.g., \cite[Sections~4.6.7--4.6.8]{Vovk/etal:2022book}),
where the observation space $\mathbf{Z}$ is split into a family, often finite,
of disjoint categories,
and conformal p-values are computed for each category separately.
For a relatively narrow group of conformal predictors,
object-conditional validity has been established for overlapping categories
by Jung et al.\ \cite{Jung/etal:arXiv2209}.
This has been generalized by Gibbs et al.\ \cite{Gibbs/etal:arXiv2305}
to what may be called ``fuzzy categories'':
a category becomes a nonnegative function on the object space $\mathbf{X}$,
with categories in the sense of subsets of $\mathbf{X}$
corresponding to the indicator functions on $\mathbf{X}$.
In this section we will see that the steps of moving
from disjoint to overlapping and then to ``fuzzy'' categories
become very simple and natural in conformal e-prediction.
\iftoggle{FULL}{\bluebegin
  For an artistic representation of overlapping categories,
  see Fig.~\ref{fig:Kandinsky}.
  \begin{figure}[htbp]
    \begin{center}
      \includegraphics[width=0.4\textwidth,height=0.4\textwidth]{/DnB/Work/R/Papers/OCM/26CEP/Mondrian_followers/Kandinsky_1926.jpg}
    \end{center} 
    \caption{Wassily Kandinsky. Several Circles, 1926.
      Solomon R. Guggenheim Museum, New York.
      Public domain}
    \label{fig:Kandinsky}
  \end{figure}
\blueend}{}%

Formally, we have a finite set, called \emph{taxonomy},
$\KKK$ of functions $\kappa:\mathbf{Z}\to[0,\infty)$,
and we are only interested in nonconformity e-measures $A$
such that, for all $m\in\{1,2,\dots\}$ and all $z_1,\dots,z_m\in\mathbf{Z}$,
\begin{equation}\label{eq:requirement}
  \frac
  {
    \sum_{i=1}^m
    \alpha_i \kappa(z_i)
  }
  {
    \sum_{i=1}^m
    \kappa(z_i)
  }
  \le1,
  \quad
  \forall\kappa\in\KKK,
\end{equation}
where $(\alpha_1,\dots,\alpha_m):=A(z_1,\dots,z_m)$,
and $0/0$ is interpreted as $1$ when it occurs on the left-hand side.
We will say that such an $A$ and the corresponding conformal e-predictor
are \emph{$\KKK$-conditional}.
An important special case, a \emph{group-wise taxonomy},
is where all $\kappa\in\KKK$ take values in $\{0,1\}$.
In this case we regard $\{z\in\mathbf{Z}:\kappa(z)=1\}$ as the categories.
In general, we can still regard $\kappa$ as fuzzy categories.

The requirement~\eqref{eq:requirement} formalizes the conditional validity
of the conformal e-predictor.
But we do not have efficiency inside a category $\kappa$
if \eqref{eq:requirement} holds as ``$<$'' rather than ``$=$'':
if we are interested in this category only,
we can improve confidence of predictions
without violating validity for this category
(but perhaps violating validity for other categories).
Let us say that a conformal e-predictor is \emph{$\KKK$-exact}
if it is based on a nonconformity e-measure (also called \emph{$\KKK$-exact})
satisfying \eqref{eq:requirement} with ``$\le$'' replaced by ``$=$''
for all $m$ and $z_1,\dots,z_m$.
Being exact is stronger than what is usually called admissible in decision theory,
but exact conformal e-predictors are ideal in the sense of achieving validity
without being strictly dominated inside each category.

The allowed nonconformity vectors $(\alpha_1,\dots,\alpha_m)$
for a $\KKK$-exact nonconformity e-measure
form the intersection of an affine space and the simplex
\[
  \left\{
    (\alpha_1,\dots,\alpha_{m})\in[0,1]^{m}:
    \alpha_1+\dots+\alpha_{m}=m
  \right\},
\]
and its dimension is at least $m-1-\lvert\mathcal{K}\rvert$.
This affine space can be defined as the shift by the vector $(1,\dots,1)\in\R^m$
of the orthogonal complement of the vectors
\begin{equation}\label{eq:orthogonal}
  \left(
    \kappa(z_i)
  \right)_{i=1}^{m},
  \quad
  \kappa\in\KKK.
\end{equation}
(Equivalently, we can define
the allowed nonconformity vectors $(\alpha_1,\dots,\alpha_m)$
as the intersection of the nonnegative orthant in $\R^m$
and the orthogonal complement of the vectors \eqref{eq:orthogonal}
extended by adding $(1,\dots,1)\in\R^m$
and shifted by $(1,\dots,1)\in\R^m$.)

The geometric picture given in the previous paragraph
makes the design of $\KKK$-exact conditional predictors very easy;
it is just a matter of picking a point
with nonnegative coordinates in a simply described affine space.
Even without the requirement of being $\KKK$-exact,
it is a matter of picking a point in a simply described polytope.

Validity results are easy to state for group-wise taxonomies.
(Remember that group-wise taxonomies allow overlapping categories,
and so cover many more applications
as compared with Mondrian conformal prediction.)
First let us state the result in the space domain
generalizing Proposition~\ref{prop:space}.

\begin{proposition}\label{prop:conditional-space}
  Let $\KKK$ be a group-wise taxonomy and $f$ be a $\KKK$-exact conformal e-predictor.
  For any $n$ and any $\kappa\in\KKK$,
  if $Z_1,\dots,Z_n,Z$ are IID (or exchangeable),
  \begin{equation}\label{eq:conditional-space}
    \E
    \left(
      f(Z_1,\dots,Z_n,Z)\mid\kappa(Z)=1
    \right) = 1
    \quad
    \text{a.s.}
  \end{equation}
\end{proposition}

\begin{proof}
  Let us fix $\kappa\in\KKK$ and check
  that \eqref{eq:conditional-space} is true even conditionally
  on the $\sigma$-algebra $\FFF$ generated by $\kappa(Z_1),\dots,\kappa(Z_n),\kappa(Z)$,
  by all $Z_i$, $i=1,\dots,n$, with $\kappa(Z_i)=0$,
  and by the multiset $B$ consisting of $Z$
  and all $Z_i$, $i=1,\dots,n$, with $\kappa(Z_i)=1$.
  Since, conditionally on $\FFF$ and inside the event $\kappa(Z)=1$,
  all orderings of $B$ are equiprobable (almost surely),
  the conditional expectation of $f(Z_1,\dots,Z_n,Z)$ is 1 given $\FFF$ and $\kappa(Z)=1$,
  which implies \eqref{eq:conditional-space}.
\end{proof}

Now we state a validity result in the time domain.

\begin{proposition}\label{prop:conditional-time}
  Let $\KKK$ be a group-wise taxonomy.
  Suppose the observations $Z_n$, $n=1,2,\dots$, are IID.
  Then, in the online prediction protocol,
  \begin{equation}\label{eq:conditional-time}
    \sum_{n=1}^{\infty} 1_{\{\kappa(Z_n)=1\}} = \infty
    \Longrightarrow
    \lim_{N\to\infty}
    \frac
    {\sum_{n=1}^N E_n 1_{\{\kappa(Z_n)=1\}}}
    {\sum_{n=1}^N 1_{\{\kappa(Z_n)=1\}}}
    =
    1
    \quad\text{a.s.}
  \end{equation}
  for a bounded $\KKK$-exact conformal e-predictor
  producing conformal e-variables $E_n$.
\end{proposition}

\noindent
The event $A\Rightarrow B$ in \eqref{eq:conditional-time}
is defined, as usual, as the union of $B$ and the complement of $A$;
therefore, $A\Rightarrow B$ holds almost surely
if the event $A\setminus B$ is null.

\begin{proof}[Proof of Proposition~\ref{prop:conditional-time}]
  By the strong law of large numbers,
  the antecedent of~\eqref{eq:conditional-time} holds with probability 0 or 1,
  depending on whether $\P(\kappa(Z)=1)=0$ or $\P(\kappa(Z)=1)>0$,
  where $Z$ is any of the $Z_n$.
  In the case $\P(\kappa(Z)=1)=0$, \eqref{eq:conditional-time} holds vacuously,
  so let us assume $\P(\kappa(Z)=1)>0$.
  Our goal is to prove that the consequent of~\eqref{eq:conditional-time} holds almost surely.

  By the strong law of large numbers,
  \begin{equation}\label{eq:denominator}
    \sum_{n=1}^N 1_{\{\kappa(Z_n)=1\}}
    \sim
    \P(\kappa(Z)=1) N
    \quad\text{a.s.},
  \end{equation}
  as $N\to\infty$.
  As in the proof of Proposition~\ref{prop:time},
  we can obtain
  \begin{equation}\label{eq:numerator}
    \sum_{n=1}^N E_n 1_{\{\kappa(Z_n)=1\}}
    \sim
    \P(\kappa(Z)=1) N
    \quad\text{a.s.}
  \end{equation}
  Combining \eqref{eq:denominator} and \eqref{eq:numerator}
  gives the consequent of~\eqref{eq:conditional-time} holding almost surely.
\end{proof}

\begin{remark}
  The proofs of Propositions~\ref{prop:conditional-space} and~\ref{prop:conditional-time}
  show that their assumptions (exchangeability or being IID)
  can be weakened, similarly to the case of Mondrian conformal prediction
  \cite[Sect.~11.3.6]{Vovk/etal:2022book}.
  On the other hand,
  the conclusion of Proposition~\ref{prop:conditional-time}
  can be strengthened to give an iterated-logarithm result
  along the lines of Proposition~\ref{prop:time-LIL}.
\end{remark}

\section{Split conformal e-predictors}
\label{sec:split}

The versions of conformal e-predictors discussed so far
are computationally feasible for a large training set
only for a narrow class of nonconformity e-measures.
The ideas of split conformal e-predictors,
discussed in this section,
and cross-conformal e-predictors, discussed in the next one,
make it possible to extend greatly the practical applicability
of conformal e-prediction.

Let us fix a measurable space $\Sigma$ (a \emph{summary space}).
A \emph{$\Sigma$-valued split nonconformity measure} is a measurable function
$A:\mathbf{Z}^+\to\Sigma$.
Intuitively, $A(z_1,\dots,z_m,z)$ encodes how well $z$ conforms to $z_1,\dots,z_m$.
A \emph{normalizing transformation} $N:\Sigma^+\to[0,\infty)^+$
is an equivariant measurable function that maps every non-empty finite sequence
$(\sigma_1,\dots,\sigma_m)$ of elements of $\Sigma$
to a finite sequence $(\alpha_1,\dots,\alpha_m)$ of the same length of nonnegative numbers
whose average is at most 1 (i.e., satisfying \eqref{eq:1}).
It is \emph{admissible} if \eqref{eq:1} holds with ``$=$''.

To apply split conformal e-prediction to a training set $z_1,\dots,z_n$,
we split it into two parts,
the \emph{training set proper} $z_1,\dots,z_{n-c}$ and the \emph{calibration set} $z_{n-c+1},\dots,z_n$.
For a new object $x$ and a potential label $y$ for it,
we set
\begin{equation}\label{eq:split}
  f(z_1,\dots,z_n,x,y)
  :=
  \alpha^y
\end{equation}
where $\alpha^y$ is defined using the following steps:
\begin{align*}
  \sigma_i &:= A(z_1,\dots,z_{n-c},z_{n-c+i}), \quad i=1,\dots,c,\\
  \sigma^y &:= A(z_1,\dots,z_{n-c},(x,y)),\\
  (\alpha_1^y,\dots,\alpha_c^y,\alpha^y) &:= N(\sigma_1,\dots,\sigma_c,\sigma^y).
\end{align*}

For many choices of $A$ and $N$, the split conformal e-predictor \eqref{eq:split}
will be computationally efficient;
this is the case when:
\begin{enumerate}
\item\label{it:rule-1}
  Processing the training set proper only once,
  we can find an easily computable rule transforming $z$ into $A(z_1,\dots,z_{n-c},z)$.
\item 
  The normalizing transformation $N$ is easily computable.
\end{enumerate}
To give examples of such easily computable $A$ and $N$,
suppose we have chosen a suitable learning architecture,
such as neural networks,
and a way of training it.
In the case of pattern recognition,
a trained neural network implements a function
$F:\mathbf{X}\to\mathbf{P}(\mathbf{Y})$,
where $\mathbf{P}(\mathbf{Y})$ is the set of all probability measures on $\mathbf{Y}$,
assumed finite and equipped with the discrete $\sigma$-algebra.
Given a test object $x\in\mathbf{X}$,
this neural network outputs a probability forecast $F(x)\in\mathbf{P}(\mathbf{Y})$ for its label:
the true label of $x$ is $y$ with probability $F(x)(\{y\})$.
An example of an easily computable (at the prediction stage) rule $A$ is
\begin{equation}\label{eq:example}
  A(z_1,\dots,z_{n-c},(x,y))
  :=
  \frac{1}{F_{z_1,\dots,z_{n-c}}(x)(\{y\})},
\end{equation}
where $F_{z_1,\dots,z_{n-c}}:\mathbf{X}\to\mathbf{P}(\mathbf{Y})$
is the neural network found from $z_1,\dots,z_{n-c}$ as training set.
(Training might take a long time, but applying the rule to a new object $x$
is typically quick.)
An example of an easily computable normalizing transformation is
\[
  (\sigma_1,\dots,\sigma_m)
  \mapsto
  \frac{m}{\sum_{i=1}^m\sigma_i}
  (\sigma_1,\dots,\sigma_m)
\]
(cf.\ \eqref{eq:normalizing}),
where the summary space is supposed to be $\Sigma\subseteq[0,\infty)$.

Proposition~\ref{prop:space}, our statement of validity,
continues to hold for split conformal e-predictors.

\begin{proposition}\label{prop:split-space}
  For any split conformal e-predictor $f$ and any $n$,
  if $Z_1,\dots,Z_n,(X,Y)$ are exchangeable,
  we have \eqref{eq:space}
  (with ``$=$'' if $f$ is admissible).
\end{proposition}

\begin{proof}
  It suffices to notice that \eqref{eq:space} holds even conditionally
  on knowing the observations $Z_1,\dots,Z_{n-c}$
  and the multiset $\lbag Z_{n-c+1},\dots,Z_{n},(X,Y)\rbag$
  since all orderings of $\lbag Z_{n-c+1},\dots,Z_{n},(X,Y)\rbag$
  are equiprobable almost surely.
\end{proof}

\section{Cross-conformal e-predictors}
\label{sec:cross}

Split conformal e-predictors are often computationally efficient,
but their predictive efficiency
(to be discussed in detail in the next section)
may suffer as compared
with ``full'' conformal e-predictors discussed in Sections~\ref{sec:CeP}--\ref{sec:conditional},
since the latter may be said to use the full training set
both as training set proper and as calibration set.
The idea behind cross-conformal e-prediction is to combine several split conformal predictors
in order to achieve better predictive efficiency.

A $\Sigma$-valued split nonconformity measure $A$ is a \emph{$\Sigma$-valued cross-nonconformity  measure}
if $A(z_1,\dots,z_m,z)$ does not depend on the order of its first $m$ arguments.
Given such an $A$ and a normalizing transformation $N$,
the corresponding \emph{cross-conformal e-predictor} is defined as follows.
The training sequence $z_1,\dots,z_n$ is randomly split into $K$ non-empty multisets (\emph{folds})
$z_{S_k}$, $k=1,\dots,K$, of equal (or as equal as possible) sizes $\left|S_k\right|$,
where $K\in\{2,3,\dots\}$ is a parameter of the algorithm,
$(S_1,\dots,S_K)$ is a partition of the index set $\{1,\dots,n\}$,
and $z_{S_k}$ consists of all $z_i$, $i\in S_k$.
For each $k\in\{1,\dots,K\}$ and each potential label $y\in\mathbf{Y}$ of the new object $x$,
find the output $\alpha_k$ of the split conformal e-predictor (based on $A$ and $N$)
on the new object $x$ and its postulated label $y$
with $z_{S_{-k}}$ as training set proper and $z_{S_k}$ as calibration set,
where $S_{-k}:=\cup_{j\ne k}S_j=\{1,\dots,n\}\setminus S_k$ is the complement to $S_k$
(and so $z_{S_{-k}}$ is the complement to the fold $z_{S_k}$).
The corresponding cross-conformal e-predictor is defined by
\begin{equation*} 
  f(z_1,\dots,z_n,x,y)
  :=
  \frac{1}{K}
  \sum_{k=1}^K
  \alpha_k.
\end{equation*}
(A slight modification, still provably valid, of this definition
is where the arithmetic mean is replaced by the weighted mean
with the weights proportional to the sizes $\left|S_k\right|$ of the folds.)

Proposition~\ref{prop:space} still holds for cross-conformal e-predictors.

\begin{proposition} 
  For any cross-conformal e-predictor $f$ and any $n$,
  if $Z_1,\dots,Z_n,(X,Y)$ are exchangeable,
  we have \eqref{eq:space}
  (with ``$=$'' if $f$ is admissible).
\end{proposition}

\begin{proof}
  This follows from Proposition~\ref{prop:split-space}
  and the arithmetic mean of e-variables being an e-variable
  (this is obvious and discussed in detail in \cite[Sect.~3]{Vovk/Wang:2021}).
\end{proof}

\begin{remark}
  To compare informally the outputs of cross-conformal predictors
  \cite[Sect.~4.4]{Vovk/etal:2022book}
  and cross-conformal e-predictors,
  we can use the rough transformation discussed in \cite[Remark~2.3]{Vovk/Wang:2021}:
  a p-value of $p$ roughly corresponds to an e-value of $1/p$.
  Under this transformation,
  the arithmetic average of e-values corresponds to the harmonic average of p-values,
  and the harmonic average is always less than or equal to the arithmetic average
  \cite[Theorem 16]{Hardy/etal:1952}.
  This suggests that cross-conformal e-prediction produces better results
  than cross-conformal prediction does.
  In the opposite direction,
  the arithmetic average of p-values corresponds to the harmonic average of e-values,
  which again suggests that cross-conformal e-prediction produces better results.
\end{remark}

\begin{remark}
  It is easy to combine the ideas of this section and Sect.~\ref {sec:conditional}
  to design conditional cross-conformal e-predictors
  (or to combine Sections~\ref{sec:split} and~\ref{sec:conditional}),
  but we stick to the simplest cases.
\end{remark}

\begin{remark}
  Proposition~\ref{prop:time} continues to hold for cross-conformal e-predictors and,
  therefore, it gives its time-wise property of validity in the online mode.
  However, the online mode entails a massive loss of their computational efficiency.
  Intuitively, using cross-conformal e-prediction in the online mode
  defeats the purpose of cross-conformal prediction:
  once we process $x_1,y_1,\dots,x_n,y_n$,
  we would like to apply the rule that we have found
  (see item \ref{it:rule-1} on p.~\pageref{it:rule-1})
  to a large number of new objects rather than just one.
  There are more complicated settings of ``weak teachers''
  (along the lines of \cite[Sect.~3.3]{Vovk/etal:2022book})
  that combine cross-conformal e-prediction with time-wise validity in useful ways,
  but we will not discuss them further in this paper.
\end{remark}

\section{Predictive efficiency of conformal e-predictors}
\label{sec:efficiency}

So far we have concentrated on the validity of conformal e-predictors.
A valid e\-/pre\-dic\-tor is not allowed to output consistently
large e-values for the true labels;
namely, the expectation of the e-value for the true label should not exceed 1.
On the other hand, for the other labels (we will call them \emph{false labels}),
we would like their e-values to be as large as possible,
and this (informal) desideratum is known as \emph{efficiency}
(or predictive efficiency,
if there is a risk of confusion with computational efficiency).
The topic of this section is ways of measuring the efficiency for conformal e-prediction.

We developed suitable criteria of efficiency for conformal prediction
in \cite[Sect.~3.1]{Vovk/etal:2022book}
(whose notation we will use in this section, except that $e$ will stand for e-values).
The idea is that a criterion of efficiency cannot be regarded as suitable
if in the limiting case of infinite training and test sets it leads
to very unnatural optimal nonconformity measures.
An example of such an unsuitable criterion of efficiency for conformal prediction
is the use of average confidence and credibility,
as defined in the first edition of \cite{Vovk/etal:2022book}
(analogously to our definition in Sect.~\ref{sec:CeP} above)
and analysed in \cite[Sections 3.1.6--3.1.7]{Vovk/etal:2022book}
(where extremely awkward features of this criterion are discussed).
This section proposes natural criteria of efficiency
for conformal e-prediction
that lead to natural nonconformity e-measures.

Suppose we have a training set $z_1,\dots,z_n\in\mathbf{Z}$,
and we are given a test set $z_{n+1},\dots,z_{n+k}$,
where $z_i=(x_i,y_i)$ for all $i$.
Let $e_i^{y}:=f(z_1,\dots,z_n,x_i,y)$ be the e-value
computed by a given conformal e-predictor $f$
for a postulated label $y$ for a test object $x_i$,
$i\in\{n+1,\dots,n+k\}$.
The average sum
\begin{equation}\label{eq:observed}
  \frac1k
  \sum_{i=n+1}^{n+k}
  \sum_{y\ne y_i}
  \ln e_i^{y}
\end{equation}
of the log e-values for the false test labels
may serve as a measure of the predictive efficiency of $f$ on the test set;
we would like it to be as large as possible.
Let us call it the \emph{observed log criterion} of efficiency
(the modifier ``observed'' will be discussed in Remark~\ref{rem:prior}).
The expression~\eqref{eq:observed} is natural insofar
as averaging logarithms of e-values is ubiquitous in numerous contexts:
see, e.g.,
the five reasons given in \cite[(7) and Sect.~3.1]{Grunwald/etal:2024}.

Let us find the optimal nonconformity e-measure for the observed log criterion
in the ``idealised setting''
akin to the one considered in \cite[Sect.~3.1.4]{Vovk/etal:2022book} in the case of conformal prediction.
For that it will be convenient to modify slightly the definition of a nonconformity e-measure.

An equivalent definition of a \emph{nonconformity e-measure} $A$
is as a measurable function mapping a nonempty multiset $\lbag z_1,\dots,z_m\rbag$,
for any $m\in\{1,2,\dots\}$,
and its element $z_i$, $i\in\{1,\dots,m\}$, to a nonnegative number
satisfying
\[
  \sum_{i=1}^m
  A(\lbag z_1,\dots,z_m\rbag,z_i)
  \le
  1
\]
for all $m\in\{1,2,\dots\}$ and all multisets $\lbag z_1,\dots,z_m\rbag$ of size $m$.
The corresponding conformal e-predictor is then defined by
\begin{equation}\label{eq:CeP-2}
  f(z_1,\dots,z_n,x,y)
  :=
  A(\lbag z_1,\dots,z_n,(x,y)\rbag,(x,y)).
\end{equation}
It is clear that this definition is equivalent to our original definition \eqref{eq:CeP-1}.
\iftoggle{FULL}{\bluebegin
  An alternative definition is \emph{deleted},
  as explained in \cite[Sect.~2.9.2]{Vovk/etal:2022book}.
  In the deleted version,
  the definition of conformal e-predictors is particularly simple:
  \[
    f(z_1,\dots,z_n,x,y)
    :=
    A(\lbag z_1,\dots,z_n\rbag,(x,y)).
  \]
\blueend}{}%

An \emph{idealised nonconformity e-measure} is a function
$A:\mathbf{P}(\mathbf{Z})\times\mathbf{Z}\to[0,\infty)$
such that $A(Q,z)$ is measurable in $z\in\mathbf{Z}$ and $\int A(Q,z)Q(\d z)\le1$
for any $Q\in\mathbf{P}(\mathbf{Z})$.
The intuition behind this definition is that $Q$ represents an infinite training set,
which becomes, once the order of its elements is forgotten,
the probability distribution of the data.
The corresponding \emph{idealised conformal e-predictor} is
$f(Q,x,y):=A(Q,x,y)$
according to \eqref{eq:CeP-2}
(adding another observation to an infinite training set does not change anything);
the difference between a nonconformity e-measure and the corresponding conformal e-predictor
disappears in the limit.

As in \cite[Sect.~3.1.5]{Vovk/etal:2022book},
we assume that the object space $\mathbf{X}$ and the label space $\mathbf{Y}$ are finite
(the latter simply means that we are interested in pattern recognition).
We identify a probability measure $Q$ on a finite set $A$ (such as $\mathbf{Y}$)
with a function mapping $a\in A$ to $Q(\{a\})$,
thus often omitting the curly braces in expressions such as $Q(\{a\})$.
If $Q$ is a probability measure on $\mathbf{Z}=\mathbf{X}\times\mathbf{Y}$,
let $Q_{\mathbf{X}}\in\mathbf{P}(\mathbf{X})$ be its marginal probability measure on $\mathbf{X}$,
and let $Q_x\in\mathbf{P}(\mathbf{Y})$, $x\in\mathbf{X}$,
be the conditional probability measure on $\mathbf{Y}$ given $x$:
\[
  Q_{\mathbf{X}}(x)
  :=
  Q(\{x\}\times\mathbf{Y}),
  \quad
  Q_x(y)
  :=
  \frac{Q(x,y)}{Q_{\mathbf{X}}(x)}.
\]

Let us fix the true data-generating probability measure $P\in\mathbf{P}(\mathbf{Z})$,
representing a given infinitely large training set.
For simplicity, let us assume that $P(z)>0$ for all $z\in\mathbf{Z}$;
this makes $P_x(y)$ well-defined and positive for all $x\in\mathbf{X}$ and $y\in\mathbf{Y}$.
To formalize the test set being infinitely large as well,
we replace the problem of maximizing \eqref{eq:observed}
by the idealised optimization problem
\begin{equation}\label{eq:observed-ideal}
  \int_{\mathbf{Z}}
  \sum_{y'\in\mathbf{Y}\setminus\{y\}}
  \ln f(P,x,y') \,
  P(\d(x,y))
  \to
  \max,
\end{equation}
$f=A$ ranging over the conformal e-predictors
(equivalently, over the nonconformity e-measures).
This corresponds to letting $k\to\infty$ in \eqref{eq:observed}.

\begin{proposition}\label{prop:observed}
  The optimal nonconformity e-measure $A$
  under the observed log criterion \eqref{eq:observed-ideal}
  is given by the odds
  \begin{equation}\label{eq:odds}
    A(P,x,y)
    :=
    \frac{1}{\lvert\mathbf{Y}\rvert-1}
    \frac{1-P_x(y)}{P_x(y)}
  \end{equation}
  against the true label being $y$ conditional on the object $x$.
\end{proposition}

\begin{proof}
  The rest of this section will be mainly devoted
  to the proof of Proposition~\ref{prop:observed}.
  First let us check that \eqref{eq:odds} is indeed a nonconformity e-measure.
  It is even true that its average is 1 over each $P_x$:
  \begin{equation*}
    \int A(P,x,y) \, P_x(\d y)
    =
    \frac{1}{\lvert\mathbf{Y}\rvert-1}
    \sum_{y\in\mathbf{Y}}
    (1-P_x(y))
    =
    1.
  \end{equation*}

  We start from the case where we have no objects,
  corresponding to $\lvert\mathbf{X}\rvert=1$
  (the only object does not carry any information).
  This case is not only a gentle introduction to the general case,
  but also corresponds to the situation where our prediction
  is fully conditional on the object $x$
  (remember that earlier we assumed $\lvert\mathbf{X}\rvert<\infty$,
  and so the full conditioning is feasible
  for an infinite training set).
  Now we can regard $P$ to be a probability measure on the label space $\mathbf{Y}$,
  and our goal is to find the optimal e-variable $E=Q/P$,
  where $Q$ is an alternative probability measure on $\mathbf{Y}$.

  In the case $\lvert\mathbf{X}\rvert=1$,
  the optimization problem \eqref{eq:observed-ideal} is
  \[
    \int\ln\frac{Q}{P}\dd1
    -
    \int\ln\frac{Q}{P}\dd P
    \to\max,
  \]
  where $1$ is the counting measure on $\mathbf{Y}$
  ($1(y):=1$ for all $y\in\mathbf{Y}$).
  This optimization problem
  \iftoggle{FULL}{\bluebegin is equivalent to
    \[
      \int\ln\frac{Q}{P}\dd(1-P)
      \to\max,
    \]
    which in turn
  \blueend}{}%
  is equivalent to
  \[
    \int\ln Q\dd(1-P)
    \to\max,
  \]
  which gives $Q(y)\propto 1-P(y)$.
  So the optimal e-values are
  \[
    E(y)
    \propto
    \frac{1-P(y)}{P(y)};
  \]
  i.e., $E(y)$ is proportional to the odds against observing label $y$.
  The full expression is
  \[
    E(y)
    =
    \frac{1}{\lvert\mathbf{Y}\rvert-1}
    \frac{1-P(y)}{P(y)},
  \]
  which agrees with \eqref{eq:odds}.

  Now let us get rid of the assumption $\lvert\mathbf{X}\rvert=1$.
  We will apply the result of the previous paragraph to each $P_x$, $x\in\mathbf{X}$.
  To find an explicit expression for the optimal e-variable
  $E:\mathbf{Z}\to[0,\infty)$,
  we split it into $E_x:\mathbf{Y}\to[0,\infty)$, $x\in\mathbf{X}$,
  defined by $E_x(y):=E(x,y)$.
  Our optimization problem is
  \begin{equation}\label{eq:quality}
    \int_{\mathbf{X}}
    \int_{\mathbf{Y}}
    \ln E_x
    \dd(1-P_x)
    P_{\mathbf{X}}(\d x)
    \to
    \max
  \end{equation}
  subject to the constraint
  \begin{equation}\label{eq:constraint}
    \int_{\mathbf{X}}
    \int_{\mathbf{Y}}
    E_x
    \dd P_x
    P_{\mathbf{X}}(\d x)
    \le
    1;
  \end{equation}
  without loss of generality, we replace ``$\le$'' by ``$=$'' in \eqref{eq:constraint}.
  Now the constraint \eqref{eq:constraint} can be rewritten as
  \[
    \int_{\mathbf{Y}}
    E_x
    \dd P_x
    =
    1+\gamma_x,
    \quad
    \int_{\mathbf{X}}
    \gamma_x
    P_{\mathbf{X}}(\d x)
    =
    0,
  \]
  where the new variables $\gamma_x$ take values in $[-1,\infty)$.
  By the result of the previous paragraph,
  \[
    E_x
    =
    (1+\gamma_x)
    \tilde E_x,
  \]
  where
  \[
    \tilde E_x(y)
    =
    \frac{1}{\lvert\mathbf{Y}\rvert-1}
    \frac{1-P_x(y)}{P_x(y)}
  \]
  is the normalized version of $E_x$.
  Maximizing the overall objective function in~\eqref{eq:quality} can be rewritten as
  \begin{equation*}
    \int_{\mathbf{X}}
    \int_{\mathbf{Y}}
    \ln\tilde E_x
    \dd(1-P_x)
    P_{\mathbf{X}}(\d x)
    +
    (\lvert\mathbf{Y}\rvert-1)
    \int_{\mathbf{X}}
    \ln(1+\gamma_x)
    P_{\mathbf{X}}(\d x)
    \to
    \max,
  \end{equation*}
  and so our optimization problem reduces to
  \begin{equation}\label{eq:objective}
    \int_{\mathbf{X}}
    \ln(1+\gamma_x)
    P_{\mathbf{X}}(\d x)
    \to
    \max
  \end{equation}
  under the constraint
  \[
    \int_{\mathbf{X}}
    \gamma_x
    P_{\mathbf{X}}(\d x)
    =
    0.
  \]
  By Jensen's inequality,
  the max in \eqref{eq:objective} is 0,
  and it is attained for $\gamma_x=0$.
  This completes the proof of Proposition~\ref{prop:observed}.
\end{proof}

\begin{remark}\label{rem:prior}
  In the terminology of \cite[Sect.~3.1]{Vovk/etal:2022book},
  the criterion of efficiency \eqref{eq:observed} is ``observed''
  in that the efficiency of the conformal e-predictor on a test observation $z_i$
  is measured by an expression,
  namely $\sum_{y\ne y_i}\ln e_i^{y}$,
  that depends on the observed true label $y_i$.
  A natural alternative to \eqref{eq:observed} is
  the \emph{prior log criterion}
  \begin{equation}\label{eq:prior}
    \frac1k
    \sum_{i=n+1}^{n+k}
    \sum_{y}
    \ln e_i^{y}
  \end{equation}
  obtained by replacing $\sum_{y\ne y_i}$ by $\sum_{y}$.
  The criterion \eqref{eq:prior} is called ``prior''
  since for it the dependence on the true label disappears;
  we can compute the sum $\sum_y$ prior to observing the true label.
  Its idealised version is
  \begin{equation*} 
    \int_{\mathbf{X}}
    \sum_{y'\in\mathbf{Y}}
    \ln f(P,x,y')
    P_{\mathbf{X}}(\d x)
    \to
    \max.
  \end{equation*}
  Adapting the argument given above to this idealised criterion
  (and slightly simplifying the argument),
  we can see that the optimal nonconformity e-measure under this criterion is
  \[
    A(P,x,y)
    :=
    \frac{1}{\lvert\mathbf{Y}\rvert P_x(y)}.
  \]
  (We have already used this nonconformity e-measure with $P_x$ replaced by its estimate:
  see \eqref{eq:example} above.)
  Notice that while in conformal prediction
  suitable (``conditionally proper'') observed and prior criteria of efficiency
  lead to the same optimal conformity measures \cite[Theorem 3.1]{Vovk/etal:2022book},
  in the case of conformal e-prediction the optimal nonconformity e-measures are different,
  albeit typically very close for a large label space~$\mathbf{Y}$.
  \iftoggle{FULL}{\bluebegin

    Let me prove the statement of optimality for the prior criterion of efficiency.
    First let us consider the fully conditional case.
    In the case of the prior criterion~\eqref{eq:prior},
    the optimization problem
    \[
      \sum\ln\frac{Q}{P}
      \to\max
    \]
    is equivalent to
    \[
      \sum\ln Q
      \to\max,
    \]
    which gives $Q(y)=1/\lvert\mathbf{Y}\rvert$.
    So the optimal E-values are
    \[
      E(y)
      :=
      \frac{1}{\lvert\mathbf{Y}\rvert P(y)}.
    \]
    Etc.
  \blueend}{}%
\end{remark}

\iftoggle{FULL}{\bluebegin
  \section{No conformal e-predictive distributions}
  \label{sec:no-CPD}

  As discussed in \cite[Part~II, especially Chap.~7]{Vovk/etal:2022book},
  a great advantage of conformal prediction is that in regression problems
  it allows us to produce ``conformal predictive distributions'',
  which are predictive distributions that are automatically well-calibrated
  (under exchangeability).
  This seems to be impossible in the context of e-values.
\blueend}{}%

\iftoggle{FULL}{\bluebegin
  \section{Online compression models}
  \label{sec:OCM}

  Define conformal e-prediction for OCM following \cite[Part IV]{Vovk/etal:2022book}.

  Discuss the important special case of the Gaussian model
  (the classical case at the birth of modern theoretical statistics),
  perhaps following \cite{Vovk/Wang:2023}
  (in whose Sect.~2 we use the normalized $\lvert x\rvert^d$
  as toy example in explanations,
  and in whose Sect.~8 we use the normalized the normalized $\lvert x\rvert^d$
  as e-variable for our empirical studies).

  The \emph{Student t-distribution} with $\nu$ degrees of freedom,
  where $\nu\in\{1,2,\dots\}$,
  is the distribution of the random variable
  \[
    \frac{\xi}{\frac{1}{\nu}\sum_{i=1}^{\nu}\xi_i},
  \]
  where $\xi,\xi_1,\dots,\xi_{\nu}$ are independent standard Gaussian random variables.
  The $d$th moment of this distribution is
  \[
    \mu_d
    :=
    \nu^{d/2}
    \frac{
      \Gamma\left(\frac{d+1}{2}\right)
      \Gamma\left(\frac{\nu-d}{2}\right)
    }{\sqrt{\pi}\Gamma(\nu/2)}
  \]
  for an even $d$
  (for an odd $d$ it is 0 by symmetry).
  For a proof, see \cite[Sect.~16.11 and Example 3.3]{Kendall:1994d}.
  (Considering even positive $d$ is sufficient for this paper,
  but the proof shows that $d$ can be any positive number
  if $\mu_d$ is understood to be the absolute moment.)

  We will be using the fact that
  \begin{equation}\label{eq:t}
    t
    :=
    \sqrt{\frac{n}{n+1}}
    \frac{z_{n}-\bar z_n}{\hat\sigma_n}
  \end{equation}
  has the Student t-distribution with $n-1$ degrees of freedom
  when $(z_1,\dots,z_n,z)$ are distributed uniformly on the \textbf{sphere}
  (as explained in \cite[Sect.~11.4.1, (11.15)]{Vovk/etal:2022book}).
  This gives us the conformal e-predictor
  \begin{equation}\label{eq:Gaussian}
    f(z_1,\dots,z_n,z)
    :=
    \frac{\lvert t\rvert^d}{\mu_d}
    =
    \frac{
      \sqrt{n\pi} \Gamma(\nu/2)}
    {((n+1)d)^{\nu/2}
      \Gamma\left(\frac{d+1}{2}\right)
      \Gamma\left(\frac{\nu-d}{2}\right)
      \hat\sigma_n^d}
    (z_{n}-\bar z_n),
  \end{equation}
  where $t$ is defined by \eqref{eq:t}.

  \begin{figure}[htbp]
    \begin{center}
      \includegraphics[width=0.48\textwidth]{direct_10.pdf}
      \quad
      \includegraphics[width=0.48\textwidth]{logarithmic_10.pdf}
    \end{center} 
    \caption{E-values and p-values for $n=10$ and for a range of $t$}\label{fig:n10}
  \end{figure}

  Figure~\ref{fig:n10} shows the Gaussian conformal e-prediction \eqref{eq:Gaussian}
  as function of $t$ for a training set of size $10$
  (so that the number of degrees of freedom is $9$).
  The thresholds shown are Jeffreys's $10^{1/2}$, $10$, $10^{3/2}$, and $100$,
  and we set $d:=4$.
  The corresponding e-values are compared with two ways of calibrating
  the standard Student--Fisher p-values into e-values.
  One is not a bona fide e-value,
  and it is sometimes referred to as the ``Vovk--Sellke bound''
  \cite{Bayarri/etal:2016,Shafer/Vovk:2019},
  \[
    \VS(p)
    :=
    \begin{cases}
      -\frac{1}{e p\ln p} & \text{if $p\le1/e$}\\
      1 & \text{otherwise};
    \end{cases}
  \]
  it is an upper bound for a very natural class of calibrator.
  Shafer's calibrator \cite{Shafer:2021}
  \[
    \Sh(p)
    :=
    p^{-1/2}-1
  \]
  gives proper e-values,
  and we apply it to the Student--Fisher p-values.
  It is interesting that our e-values
  dominate both Vovk--Sellke bound and Shafer's calibrator
  in Jeffreys's range $[10^{1/2},10^2]$.

  \begin{figure}[htbp]
    \begin{center}
      \includegraphics[width=0.48\textwidth]{direct_100.pdf}
      \quad
      \includegraphics[width=0.48\textwidth]{logarithmic_100.pdf}
    \end{center} 
    \caption{E-values and p-values for $n=100$ and for a range of $t$}\label{fig:n100}
  \end{figure}

  Figure~\ref{fig:n100} is the analogue of Fig.~\ref{fig:n10}
  for $n=100$ (and so 99 degrees of freedom).
  It is interesting that $d=4$ is now does not perform well
  in Jeffreys's range $[10^{1/2},10^2]$,
  and so we take $d=6$.
  Shafer's calibrator does not seem to perform well
  for both $n=10$ and $n=100$
  as compared with the Vovk--Sellke bound,
  but we should keep in mind that the letter is overoptimistic.

  \section{Simulation studies}

  A natural conformity e-measure for use in simulation studies:
  Bayesian conformity e-measure.

  Useful rule of thumb \cite{Jeffreys:1961}:
  weak, substantial, strong, very strong, decisive prediction sets.

  Comparison with cross-conformal prediction using the Vovk--Sellke bound in simulation studies.
\blueend}{}%

\section{Conclusion}
\label{sec:conclusion}

In this paper we have discussed three strengths of conformal prediction
(even if two of them briefly):
\begin{enumerate}
\item\label{it:1}
  As set predictors,
  conformal predictors possess a property of validity
  that is both reasonably strong and very simple:
  their probability of error is bounded by a prespecified constant.
  This advantage is lost in conformal e-prediction
  (the probability of error being bounded by a prespecified constant
  becomes a very weak property of validity for conformal e-predictors,
  and stronger properties of validity,
  such as the one given at the end of \ref{app:BB},
  are somewhat less intuitive).
\item\label{it:2}
  In the online prediction protocol,
  conformal predictors as set predictors make errors at different steps independently.
  Conformal e-predictors satisfy a weakened version of this property,
  as discussed in Sect.~\ref{sec:validity}.
\item\label{it:4}
  In the case of regression,
  conformal prediction can be used for producing conformal predictive distributions,
  and so conformal predictors can also be used as probabilistic predictors.
  In this role, conformal predictors are automatically well-calibrated.
  This advantage is lost for conformal e-prediction.
\end{enumerate}
Two of these strengths, \ref{it:1} and \ref{it:4}, appear to be clear advantages
of conformal prediction over conformal e-prediction.
For strength \ref{it:2}, this is less obvious,
but the picture for conformal prediction still appears simpler and nicer.

Conformal e-prediction has at least two advantages of its own:
\begin{itemize}
\item
  Designing conditional conformal e-predictors is much easier
  and using e-values adds flexibility,
  especially as compared with Mondrian conformal predictors;
  see Sect.~\ref{sec:conditional}.
\item
  Cross-conformal e-predictors are provably valid,
  unlike cross-conformal predictors,
  as discussed in Sect.~\ref{sec:cross}.
\end{itemize}
Looking for further advantages of conformal e-prediction
is an interesting direction of further research.

\iftoggle{FULL}{\bluebegin
  Open problems:
  \begin{itemize}
  \item
    Are there interesting analogues of the criteria of efficiency
    discussed in Sect.~\ref{sec:efficiency}
    for other online compression models (Sect.~\ref{sec:OCM})?
  \end{itemize}

  Can I implement the Burnaev--Wasserman programme for conformal e-prediction?
  (With a Bayesian soft model.)
  Perhaps in the case of object-conditional conformal e-predictors
  it will lead to the RIP (reverse information projection),
  which can be solved using Lagrange multipliers.
\blueend}{}%

\subsection*{Acknowledgments}

Many thanks to Aaditya Ramdas for pointing out deficiencies
in presentation in an earlier version
and to Alex Balinsky and Ilia Nouretdinov for useful discussions.
The anonymous reviewers' comments helped greatly
in improving the presentation further.
This research has been partially supported by Astra Zeneca, Stena Line, and Mitie.


\iftoggle{JOURNAL}{%
  \bibliographystyle{elsarticle-num}
}{}%
\iftoggle{CONF}{%
  \bibliographystyle{splncs04}
}{}%
\iftoggle{TR}{%
  \bibliographystyle{plain}
}{}%
\bibliography{local,%
  /doc/work/r/bib/AIT/AIT,%
  /doc/work/r/bib/general/general,%
  /doc/work/r/bib/math/math,%
  /doc/work/r/bib/prob/prob,%
  /doc/work/r/bib/stat/stat,%
  /doc/work/r/bib/vovk/vovk}%

\appendix
\section{Law of the iterated logarithm for e-flows}
\label{app:LIL}

The main goal of this appendix is to prove
the law of the iterated logarithm in the form of Proposition~\ref{prop:time-LIL}.
As I could not find this version of the law of the iterated logarithm in literature,
I will state it in a general form in this appendix.

Let $\FFF_n$, where $n\in\Z$ and $\Z$ stands for the set of all integer numbers,
be an increasing sequence of $\sigma$-algebras
on a given sample space $\Omega$,
\[
  \dots\subseteq\GGG_{-1}\subseteq\GGG_{0}\subseteq\GGG_{1}\subseteq\dots,
\]
and $\xi_n$, $n\in\Z$, be an adapted sequence of random variables,
meaning that each $\xi_n$ is $\GGG_n$-measurable.
Such an adapted sequence is an \emph{e-flow} if it is nonnegative and
\begin{equation}\label{eq:e-flow}
  \E(\xi_n\mid\GGG_{n-1})
  \le
  1,
  \quad
  n\in\Z.
\end{equation}
Let us say that the e-flow is \emph{exact}
if \eqref{eq:e-flow} holds with ``$=$'' in place of ``$\le$''.

The main example of an e-flow in the context of this paper is
\[
  \GGG_n
  :=
  \begin{cases}
    \FFF_{-n} & \text{if $n<0$}\\
    \FFF_{1} & \text{otherwise},
  \end{cases}
\]
where $\FFF_n$, $n=1,2,\dots$, is the \emph{exchangeable filtration}
(introduced in Sect.~\ref{sec:validity}:
$\FFF_{n}$ is generated by $\lbag Z_1,\dots,Z_{n-1}\rbag,Z_n,Z_{n+1},\dots$),
and
\[
  \xi_n
  :=
  \begin{cases}
    E_{-n} & \text{if $n<0$}\\
    1 & \text{otherwise},
  \end{cases}
\]
where $E_n$ is the $n$th conformal e-variable, $E_n:=f(Z_1,\dots,Z_n)$.
Let us call this the \emph{conformal e-flow}.
The fact that the adapted sequence $(\xi_n,\GGG_n)$
defined in this way is an e-flow is the main property of validity of conformal e-prediction.
The main part of the conformal e-flow is $(\xi_n,\GGG_n)$ for $n<0$;
the extension to all integer $n$ is vacuous.
Exact e-flows correspond to admissible conformal predictors.

There are two natural laws of the iterated logarithm for e-flows:
the \emph{forward law} describes the behaviour, as $N\to\infty$, of the sums
$\sum_{n=1}^N(\xi_n-1)$,
and the \emph{backward law} describes the behaviour of the sums
$\sum_{n=-N}^{-1}(\xi_n-1)$ .
The forward law is just a restatement
of the standard martingale law of the iterated logarithm,
since with each forward sequence $\xi_n$, $n\ge1$,
we can associate a supermartingale $X$ that carries the same information
by setting
\begin{equation*} 
  X_n
  :=
  \sum_{i=1}^n
  (\xi_i-1),
\end{equation*}
with $X_0:=0$;
remember that the defining property of being a supermartingale
is $\E(X_n\mid\FFF_{n-1})\le X_{n-1}$.
\iftoggle{FULL}{\bluebegin
  We could call $X_n$ the \emph{additive canonical supermartingale}.
  Other such processes are:
  \begin{itemize}
  \item
    the \emph{multiplicative canonical supermartingale} is
    \[
      X_n
      :=
      \prod_{i=1}^n
       \xi_i;
    \]
  \item
    the \emph{multiplicative canonical martingale} is
    \[
      X_n
      :=
      \prod_{i=1}^n
      \frac{\xi_i}{\E(\xi_i\mid\GGG_{i-1})};
    \]
  \item
    the \emph{additive canonical (test) martingale} is
    \[
      X_n
      :=
      \sum_{i=1}^n
      (\xi_i-\E(\xi_i\mid\GGG_{i-1}).
    \]
  \end{itemize}
\blueend}{}%
Applying the standard law of the iterated logarithm to the supermartingale $X_n$
gives us the following corollary.

\begin{corollary} 
  For any bounded e-flow $(\xi_n,\GGG_n)$,
  \begin{equation*}
    \limsup_{N\to\infty}
    (N\ln\ln N)^{-1/2}
    \sum_{n=1}^{N}
    (\xi_n-1)
    \le
    \sqrt{2(C-1)}
    \quad\text{a.s.},
  \end{equation*}
  where $C>1$ is an upper bound for $\xi_n$,
  and there exists an e-flow $(\xi_n,\GGG_n)$ bounded above by $C$
  (assuming $C>1$)
  such that
  \begin{equation*}
    \limsup_{N\to\infty}
    (N\ln\ln N)^{-1/2}
    \sum_{n=1}^N
    (\xi_n-1)
    =
    \sqrt{2(C-1)}
    \quad\text{a.s.}
  \end{equation*}
\end{corollary}

\noindent
(Alternatively, the proof of the backward law of the iterated logarithm
given later in this appendix will also prove the forward law,
after trivial modifications.)
The assumption $C>1$ only excludes a trivial case.

Such a reduction to the supermartingale case is impossible for the backward law;
even though there are numerous laws of the iterated logarithm for reverse martingales,
they are not applicable in our current context.
\iftoggle{FULL}{\bluebegin
  For example, in the context of exact conformal e-flow $(\xi_n,\GGG_n)$,
  the corresponding martingale $X_n$ for $n<0$
  would be completely determined by $X_{-1}$ as $X_n=\E(X_{-1}\mid\GGG_n)$.
  (As Dudley \cite[Sect.~10.6]{Dudley:2002} remarks.)
  The conformal e\-/variables $\xi_n$ would be somehow encoded in $X_{-1}$,
  which seems impossible:
  each new $\xi_n$, $n=-1,-2,\dots$, may carry new information.
  One case where an exact e-flow $(\xi_n,\GGG_n)$, $n<0$, carries the same information
  as a reverse martingale is where $\prod_{n=-\infty}^{-1}\xi_n$ converges in a suitable sense
  (in the multiplicative picture;
  in the additive picture this becomes the convergence of $\sum_{n=-\infty}^{-1}(\xi_n-1)$).
  The corresponding reverse martingale (in the case of exact e-flows) is then
  \[
    X_n
    :=
    \prod_{i=-\infty}^{n}
    \xi_i.
  \]
  (This is the way Christophe Cuny and Florence Merlev\`ede \cite{Cuny/Merlevede:2015}
  move from reverse martingale difference in the wide sense (Theorem~2.3)
  to reverse martingale difference in the narrow sense (Proposition~2.1).)

  The proof of Proposition~\ref{prop:time} also proves
  the following more general proposition.

  \begin{proposition} 
    For any bounded e-flow $(\xi_n,\GGG_n)$,
    \begin{equation*}
      \limsup_{N\to\infty}
      \frac1N
      \sum_{n=-N}^{-1}
      \xi_n
      \le
      1
      \quad\text{a.s.},
    \end{equation*}
    and for any exact e-flow $(\xi_n,\GGG_n)$,
    \begin{equation*}
      \limsup_{N\to\infty}
      \frac1N
      \sum_{n=-N}^{-1}
      \xi_n
      =
      1
      \quad\text{a.s.}
    \end{equation*}
  \end{proposition}
\blueend}{}%
Luckily, however, the standard proof of the law of the iterated logarithm
can be easily adapted to the backward case
and allows us to establish the following backward law of the iterated logarithm.

\begin{proposition}\label{prop:LIL}
  For any bounded e-flow $(\xi_n,\GGG_n)$,
  we have
  \begin{equation}\label{eq:LIL-upper} 
    \limsup_{N\to\infty}
    (N\ln\ln N)^{-1/2}
    \sum_{n=-N}^{-1}
    (\xi_n-1)
    \le
    \sqrt{2(C-1)}
    \quad\text{a.s.},
  \end{equation}
  where $C>1$ is an upper bound,
  and there exists an e-flow $(\xi_n,\GGG_n)$ bounded above by $C$
  (assuming $C>1$)
  that satisfies
  \begin{equation}\label{eq:LIL-exact} 
    \limsup_{N\to\infty}
    (N\ln\ln N)^{-1/2}
    \sum_{n=-N}^{-1}
    (\xi_n-1)
    =
    \sqrt{2(C-1)}
    \quad\text{a.s.}
  \end{equation}
\end{proposition}

\iftoggle{FULL}{\bluebegin
  Comparison with~\eqref{eq:LIL-exact} shows that the inequality~\eqref{eq:LIL-upper} is tight.

  Proposition~\ref{prop:LIL} is obviously equivalent
  to a law of the iterated logarithm for reverse martingale differences in the wide sense.
  There exists a law of the iterated logarithm for reverse martingale differences in the narrow sense
  \cite{Lin:1978},
  but it is completely different from Proposition~\ref{prop:LIL}
  (and is given in terms of the remaining variance rather than the accumulated variance).%
\blueend}{}%

\iftoggle{JOURNAL}{}{\input{suppl_1.txt}}

\iftoggle{JOURNAL}%
  {For the proof of Proposition~\ref{prop:LIL},
    see the Supplementary Material.
    Our argument given there }%
  {Our argument given above }%
proves the following more standard law of the iterated algorithm
(which we do not need in this paper).

\begin{proposition}\label{prop:standard-LIL} 
  Let $\eta_n$, $n\in\Z$, be a bounded two-sided sequence of random variables
  adapted to a filtration $\GGG_n$, $n\in\Z$.
  Suppose $\E(\eta_n\mid\GGG_{n-1})=0$ for all $n\in\Z$.
  Set
  \begin{equation}\label{eq:S-A}
    S_N
    :=
    \sum_{n=-N}^{-1}
    \eta_n
    \text{ and }
    A_N
    :=
    \sum_{n=-N}^{-1}
    \E(\eta^2_n\mid\GGG_{n-1}).
  \end{equation}
  Then
  \[
    \limsup_{N\to\infty}
    \frac{S_N}{\sqrt{2A_N\ln\ln A_N}}
    =
    1
    \quad
    \text{a.s.}
  \]
  provided $A_N\to\infty$ a.s.\ as $N\to\infty$.
\end{proposition}

\noindent
\iftoggle{JOURNAL}{}{\input{suppl_2.txt}}%
Of course, this proposition will remain true
if we replace \eqref{eq:S-A} by
\begin{equation*}
  S_N
  :=
  \sum_{n=1}^{N}
  \eta_n
  \text{ and }
  A_N
  :=
  \sum_{n=1}^{N}
  \E(\eta^2_n\mid\GGG_{n-1}),
\end{equation*}
but then it will become just a special case of the standard martingale law of the iterated logarithm.

\section{Bounding the error probability of conformal e-predictors via Markov's inequality}
\label{app:BB}

The first advantage of conformal prediction mentioned in Sect.~\ref{sec:introduction}
is that conformal predictors can be used as set predictors,
in which case their property of validity can be expressed
as a low probability of error.
In principle, this can also be done in the case of conformal e-prediction,
and has been done in \cite[Sect.~2]{Balinsky/Balinsky:2024},
but it leads to a predictor that is not admissible
in the terminology of statistical decision theory \cite[Sect.~1.3]{Wald:1950}.
This is the topic of this appendix.
For simplicity we will assume that the object and label spaces $\mathbf{X}$ and $\mathbf{Y}$
are finite.

Let us fix the size $n$ of the training set.
A \emph{set predictor} is a function
$\Gamma:\mathbf{Z}^{n}\times\mathbf{X}\to2^{\mathbf{Y}}$.
\iftoggle{FULL}{\bluebegin
  Without the assumption that $\mathbf{X}$ and $\mathbf{Y}$ are finite,
  we need to require that $\Gamma$ is measurable in the sense of the set
  \[
    \left\{
      (z_1,\dots,z_n,(x,y))\in\mathbf{Z}^{n+1}:
      y \in \Gamma(z_1,\dots,z_n,x)
    \right\}
  \]
  being measurable.
\blueend}{}%
It is \emph{$\epsilon$-valid}, where $\epsilon>0$,
if
$ \P(Y\notin\Gamma(Z_1,\dots,Z_n,X))
  \le
  \epsilon$
provided $Z_1,\dots,Z_n,(X,Y)$ are exchangeable;
in other words, if the probability of error is bounded by $\epsilon$.
An example of an $\epsilon$-valid set predictor is the \emph{conformal $\epsilon$-predictor}
\[
  \Gamma^{\epsilon}(z_1,\dots,z_n,x)
  :=
  \{y:f(z_1,\dots,z_n,x,y)>\epsilon\},
\]
where $f$ is a conformal p-predictor
\cite[Proposition~2.3]{Vovk/etal:2022book}.
Conformal $\epsilon$-predictors are just conformal predictors packaged as set predictors.

Let $\epsilon\in(0,1)$ (informally, this is our target probability of error).
The \emph{BB-predictor} \cite[Sect.~2]{Balinsky/Balinsky:2024}
associated with an admissible conformal e-predictor $f$ at significance level $1/\epsilon$
is defined as
\begin{equation}\label{eq:BB}
  \Gamma(z_1,\dots,z_n,x)
  :=
  \left\{
    y:
    f(z_1,\dots,z_n,x,y)<1/\epsilon
  \right\}.
\end{equation}
In combination with Markov's inequality,
Proposition~\ref{prop:space} implies that the BB-predictor defined in this way
is $\epsilon$-valid.

An $\epsilon$-valid set predictor $\Gamma$ is \emph{inadmissible}
if there exists an $\epsilon$-valid set predictor $\Gamma'$
such that
\begin{equation}\label{eq:domination}
  \Gamma'(z_1,\dots,z_n,x)
  \subseteq
  \Gamma(z_1,\dots,z_n,x)
\end{equation}
for all $z_1,\dots,z_n,x$
and the inclusion is strict for some $z_1,\dots,z_n,x$.
Otherwise, $\Gamma$ is \emph{admissible}.
We will say that $\Gamma'$ \emph{dominates} $\Gamma$
if \eqref{eq:domination} holds for all $z_1,\dots,z_n,x$
and that the domination is \emph{strict}
if the inclusion in \eqref{eq:domination} is strict for some $z_1,\dots,z_n,x$.


First let us check that each BB-predictor is dominated by a conformal predictor.

\begin{proposition}\iftoggle{FULL}{\label{prop:BB-1}}{}
  Let $\epsilon\in(0,1)$.
  The BB-predictor associated with an admissible conformal e-predictor $f$
  at significance level $1/\epsilon$
  is $\epsilon$-valid,
  but it is dominated by the conformal $\epsilon$-predictor based on $f$'s nonconformity e-measure.
\end{proposition}


\iftoggle{FULL}{\bluebegin
  Balinsky and Balinsky \cite[Sect.~2, Remark]{Balinsky/Balinsky:2024}
  notice that the validity of the BB-predictor does not require exchangeability
  and holds under the weaker assumption of the cycle invariance
  of the joint distribution of $(Z_1,\dots,Z_n,(X,Y))$.
  However,
  \begin{itemize}
  \item
    this is also true about the dominating conformal predictor
    constructed in the proof of Proposition~\ref{prop:BB-1},
  \item
    and the weaker assumption does not look interesting
    in the context of prediction.
  \end{itemize}
\blueend}{}%

\begin{proof} 
  The validity was checked earlier.
  Let us check that the BB-predictor is dominated
  by the conformal $\epsilon$-predictor $\Gamma^{\epsilon}$
  (based on $f$'s nonconformity e-measure as conformity measure):
  if
  \[
    \frac
      {\left|\{i=1,\dots,n+1:\alpha_i\ge\alpha_{n+1}\}\right|}
      {n+1}
    >
    \epsilon,
  \]
  then we have $\alpha_i\ge\alpha_{n+1}$ for
  at least $\lfloor(n+1)\epsilon\rfloor+1$ $\alpha_i$s,
  which implies
  \[
    \frac
      {\alpha_{n+1}}
      {\frac{1}{n+1}\sum_{i=1}^{n+1}\alpha_i}
    \le
    \frac{n+1}{\lfloor(n+1)\epsilon\rfloor+1}
    <
    \frac{1}{\epsilon}.
    \qedhere
  \]
\end{proof}

In typical cases a BB-predictor will be inadmissible,
being strictly dominated by the conformal predictor based on the same nonconformity (e-)measure.
This will be formalized in the next proposition, for which we need two definitions.

A nonconformity e-measure $A$ is \emph{generic} if it always outputs
$(\alpha_1,\dots,\alpha_{n+1})$ that are all different
(they may be only slightly different, or alternatively we can add slight randomization).
In this case we will also say that the corresponding conformal e-predictor is generic.

A nonconformity measure $A$ is \emph{$\epsilon$-categorical}
if its nonconformity scores take only two values, $0$ and $1/\epsilon$:
it only outputs $(\alpha_1,\dots,\alpha_{n+1})$ with $\alpha_i\in\{0,1/\epsilon\}$
for all $i\in\{1,\dots,n+1\}$.
Categorical nonconformity e-measures $A$ are a way of encoding set predictors:
such an $A$ represents the set predictor
\[
  \Gamma(z_1,\dots,z_n,x)
  :=
  \left\{
    y\in\mathbf{Y}:
    f(z_1,\dots,z_n,x,y)>0
  \right\},
\]
where $f$ is the corresponding conformal e-predictor.
The next proposition says that the only way to avoid inadmissibility of the BB-predictor
based on a generic nonconformity e-measure $A$
is to make $A$ very close to being $\epsilon$-categorical,
where the \emph{deviation from being $\epsilon$-categorical} is measured by
\[
  d(A)
  :=
  \max_{(z_1,\dots,z_{n+1})\in\mathbf{Z}^{n+1}}
  \left(
    \sum_{i:\alpha_i\ge1/\epsilon}
    (\alpha_i-1/\epsilon)
    +
    \sum_{i:\alpha_i<1/\epsilon}
    \alpha_i
  \right),
\]
$\alpha_1,\dots,\alpha_{n+1}$ are the nonconformity scores for $z_1,\dots,z_{n+1}$,
and $i$ ranges over $\{1,\dots,n+1\}$.
We will also say that $d(f):=d(A)$ is the deviation from being $\epsilon$-categorical
for the conformal e-predictor $f$ associated with $A$.

\begin{proposition}\label{prop:BB-2}
  Let $\epsilon\in(0,1)$.
  The BB-predictor associated with an admissible conformal e-predictor $f$
  at significance level $1/\epsilon$
  is inadmissible if $f$ is generic and $d(f)\ge1/\epsilon$.
\end{proposition}

\noindent
The lower bound of $1/\epsilon$ on the deviation $d(f)$ in Proposition~\ref{prop:BB-2}
is very small for a large size $n$ of the training set,
as the typical order of magnitude for $d(f)$ is $n$.
Therefore, BB-predictors are typically inadmissible.

\begin{proof}[Proof of Proposition~\ref{prop:BB-2}]
  In order for the BB-predictor to be admissible,
  for each $(z_1,\dots,z_{n+1})\in\mathbf{Z}^{n+1}$,
  the $\lfloor(n+1)\epsilon\rfloor$ largest $\alpha_i$
  in the corresponding $(\alpha_1,\dots,\alpha_{n+1})$
  should all be at least $1/\epsilon$.
  Therefore, the deviation from being $\epsilon$-categorical should be at most
  \[
    n+1-\frac{\lfloor(n+1)\epsilon\rfloor}{\epsilon}
    <
    n+1-\frac{(n+1)\epsilon-1}{\epsilon}
    =
    1/\epsilon.
    \qedhere
  \]
\end{proof}

The key reason for the inadmissibility of the BB-predictor in typical situations
is the strength of the validity property of conformal e-prediction.
To see this strength more clearly,
let us generalize the definition \eqref{eq:BB}
by allowing the significance level $\alpha$ to be any positive number:
\begin{equation}\label{eq:BB-general}
  \Gamma^{\alpha}(z_1,\dots,z_n,x)
  :=
  \left\{
    y:
    f(z_1,\dots,z_n,x,y)<\alpha
  \right\}.
\end{equation}
The identity
\[
  \E(E)
  =
  \int_0^{\infty}
  \P(E\ge\alpha)
  \dd\alpha
\]
allows us to state the validity property for $f$ in terms of $\Gamma^{\alpha}$ as follows:
the probability of error for $\Gamma^{\alpha}$ should integrate to at most 1 over $\alpha$.
This is much stronger than requiring the probability of error for $\Gamma^{\alpha}$
to be at most $1/\alpha$ for each $\alpha$.
The new requirement is joint rather than being a separate requirement for each $\alpha$.
\end{document}

%% file: suppl_1.txt

The proof of Proposition~\ref{prop:LIL} uses the standard basic scheme
(see, e.g., \cite[Sect.~4.4]{Shiryaev:2019}).
As a first step we fix an e-flow $(\xi_n,\GGG_n)$
and assume, without loss of generality,
that it is exact
(an easy way to see that there is no loss of generality
is to apply the idea of coupling \cite[Sect.~7.4]{Dubhashi/Panconesi:2009}).

We need an auxiliary exponential supermartingale
from Stout's proof of the law of the iterated logarithm for martingales 
\cite[Lemma~5.4.1]{Stout:1974}.

\begin{lemma} 
  Let $\eta_1,\eta_2,\dots$ be a martingale difference
  bounded above by a constant $c$, $\eta_n\le c$,
  w.r.\ to a filtration $\FFF_0,\FFF_1,\dots$.
  Set $S_n:=\sum_{i=1}^n\eta_i$.
  Let $\delta\in(0,1/c]$ be another constant.
  Then
  \begin{equation}\label{eq:T}
    T_n
    :=
    \exp
    \left(
      \delta S_n
      -
      \frac{\delta^2}{2}
      \left(
        1 + \frac{\delta c}{2}
      \right)
      \sum_{i=1}^n
      \E(\eta_i^2\mid\FFF_{i-1})
    \right)
  \end{equation}
  is a supermartingale.
\end{lemma}

In fact, Stout's lemma only assumes that $\eta_n$ is a supermartingale difference,
but we do not need this generality.
Let us derive a corollary of this lemma that will allow us to establish \eqref{eq:LIL-upper}.
First we notice that $\E(\eta_i^2\mid\FFF_{i-1})\le c$
(the largest $\E(\eta_i^2\mid\FFF_{i-1})=c$ is attained
for $\eta_i\in\{-1,c\}$ taking value $c$ with probability $1/(c+1)$);
this is spelled out in the following lemma.

\begin{lemma}\label{lem:max}
  For any $c>0$,
  $\max_{\eta}\E(\eta^2)=c$,
  $\eta$ ranging over the random variables with $\E(\eta)=0$
  and $\eta\in[-1,c]$.
  \iftoggle{FULL}{\bluebegin
    This remains true if the condition $\E(\eta)=0$ is replaced by $\E(\eta)\le0$,
    assuming $c\ge1$.%
  \blueend}{}%
\end{lemma}

\begin{proof}
  This is a special case of the Bhatia--Davis inequality \cite{Bhatia/Davis:2000}:
  see Theorem~1 there and the remark after its first proof.
  \iftoggle{FULL}{\bluebegin
    \textbf{The following is my direct proof (assuming $c\ge1$).}
    Assume, without loss of generality, that $\eta$ does not take values in $(-1,1]$
    (all such values can be mapped to $-1$;
    $\E(\eta^2)$ can only increase and $\E(\eta)\le0$ will be still satisfied).
    If $\eta$ takes a value $a$ in $(1,c)$ with a positive probability $p$,
    we can increase $a$ to $c$ reducing its probability to $(a/c)p$;
    this will increase $\E(\eta^2)$ and $\E(\eta)\le0$ will be still satisfied,
    where $\E$ may be over a subprobability measure now.
    Generalizing this argument to probability measures on $(1,c)$,
    we can assume, without loss of generality, that $\eta\in\{-1,c\}$.
    Scaling up the subprobability measure (if needed),
    let us make it a probability measure.
    If $\P(\eta=c)=p$,
    we can rewrite the condition $\E(\eta)\le0$ as $p\le\frac{1}{c+1}$,
    and so
    \[
      \E(\eta^2)
      \le
      c^2 \frac{1}{c+1}
      +
      \frac{c}{c+1}
      =
      c.
    \]
    And we can see that $\E(\eta^2)=c$ and $\E(\eta)=0$
    for $p=\frac{1}{c+1}$.%
  \blueend}{}%
\end{proof}

Since $T_n$ defined by \eqref{eq:T} is a test supermartingale,
we have, by Ville's inequality,
\begin{equation*}
  \P
  \left\{
    \max_{n=0,\dots,N}
    T_n
    \ge
    \gamma
  \right\}
  \le
  \frac{1}{\gamma}
\end{equation*}
for any $\gamma>1$ and any upper bound $N$ on $n$.
By Lemma~\ref{lem:max}
(modified to cover $\E(\eta^2\mid\FFF)$ in place of $\E(\eta^2)$),
this implies
\begin{equation}\label{eq:S}
  \P
  \left\{
    \max_{n=0,\dots,N}
    S_n
    \ge
    \frac{\delta}{2}
    \left(
      1 + \frac{\delta c}{2}
    \right)
    c N
    +
    \frac{\ln\gamma}{\delta}
  \right\}
  \le
  \frac{1}{\gamma}.
\end{equation}
The minimum over $\delta$ of the sum
\[
  \frac{\delta}{2}
  c N
  +
  \frac{\ln\gamma}{\delta}
\]
(with the term $\frac{\delta c}{2}$ in \eqref{eq:S} ignored for now)
is attained at
\begin{equation}\label{eq:delta}
  \delta
  :=
  \sqrt{2\ln\gamma/(cN)},
\end{equation}
and substituting this expression for $\delta$ into \eqref{eq:S} gives
\begin{equation}\label{eq:exponential}
  \P
  \left\{
    \max_{n=0,\dots,N}
    S_n
    \ge
    \sqrt{2c N\ln\gamma} + \frac{c\ln\gamma}{2}
  \right\}
  \le
  \frac{1}{\gamma}
\end{equation}
(the condition $\delta\le1/c$ will be satisfied
when we apply this inequality later in the proof).
The inequality~\eqref{eq:exponential} is also applicable to $-S_n$ in place of $S_n$
provided $c\ge1$,
and for any $c>0$ it becomes applicable to $-S_n$ in place of $S_n$
if we replace the second entry of $c$ in it by 1 and assume $\delta\le1$.

\iftoggle{FULL}{\bluebegin
  Let us check the statement for $-S_n$.
  Instead of \eqref{eq:S} we have
  \begin{equation}\label{eq:S}
    \P
    \left\{
      \max_{n=0,\dots,N}
      S_n
      \ge
      \frac{\delta}{2}
      \left(
        1 + \frac{\delta}{2}
      \right)
      c N
      +
      \frac{\ln\gamma}{\delta}
    \right\}
    \le
    \frac{1}{\gamma},
  \end{equation}
  assuming $\delta\le1$.
  We have the same optimal $\delta$,
  given by \eqref{eq:delta},
  and substituting this expression for $\delta$ into the modified \eqref{eq:S} gives
  \begin{equation*}
    \P
    \left\{
      \max_{n=0,\dots,N}
      S_n
      \ge
      \sqrt{2c N\ln\gamma} + \frac{\ln\gamma}{2}
    \right\}
    \le
    \frac{1}{\gamma}.
  \end{equation*}%
\blueend}{}%

\begin{proof}[Proof of Proposition~\ref{prop:LIL}]
  Let $\lambda>1$ (later we will also let $\lambda\to1$)
  and set $n_k:=\lceil\lambda^k\rceil$, $k=1,2,\dots$.
  For a given $k$,
  we will use the notation,
  for $k,n\in\{1,2,\dots\}$,
  \[
    \eta^k_n := \xi_{-n_{k}-1+n}-1
    \text{ and }
    S^k_n := \sum_{i=1}^n \eta_i,
  \]
  so that $\eta^k_n$, $n=1,2,\dots$, is a martingale difference taking values in $[-1,C-1]$
  and $S^k_n$, $n=1,2,\dots$, is a martingale.
  Set $c:=C-1$.

  Later we will choose a suitable function $\psi:\{1,2,\dots\}\to\R$;
  roughly, $\psi(n)$ will be an upper bound for $\sum_{i=-n}^{-1}(\xi_i-1)$.
  Let $A_k$ be the event that $S^k_{n_k}\ge\psi(n_k)$
  and $B_k$ be the event that $-S^k_n\ge\psi(n_k-n_{k-1})$ for some $n\in(0,n_k-n_{k-1}]$.
  Namely, we will choose $\psi$ in such a way that, for sufficiently large $k$,
  \begin{equation}\label{eq:main-1}
    \P(A_k)
    \le
    \P(S^k_n\ge\psi(n_k) \text{ for some $n\le n_{k}$})
    \le
    k^{-\lambda}
  \end{equation}
  and
  \begin{equation}\label{eq:main-2}
    \P(B_k)
    =
    \P(-S^k_n\ge\psi(n_k-n_{k-1}) \text{ for some $n\le n_k-n_{k-1}$})
    \le
    k^{-\lambda}.
  \end{equation}
  By the Borel--Cantelli lemma, as $\sum_k\P(A_k)<\infty$ and $\sum_k\P(B_k)<\infty$,
  $A_k$ and $B_k$ will hold only for finitely many $k$.

  In order for the inequalities ``${}\le k^{-\lambda}$''
  in \eqref{eq:main-1} and \eqref{eq:main-2} to hold,
  we can set, according to \eqref{eq:exponential}
  (with $-S_n$ in place of $S_n$ in the case of \eqref{eq:main-2}),
  \begin{align*}
    \psi(n_k)
    &=
    \sqrt{2c n_{k}\ln(k^{\lambda})} + \frac{c\ln(k^{\lambda})}{2}
    \sim
    \sqrt{2c n_{k} \lambda\ln k}\\
    \psi(n_k-n_{k-1})
    &=
    \sqrt{2c(n_k-n_{k-1})\ln(k^{\lambda})} + \frac{\ln(k^{\lambda})}{2}
    \sim
    \sqrt{2c(n_k-n_{k-1}) \lambda\ln k}.
  \end{align*}
  Therefore, we can choose
  $ \psi(n)
    \sim
    \sqrt{2c \lambda n\ln\ln n}$.
  The conditions $\delta\le1/c$ and $\delta\le1$ mentioned earlier
  indeed hold, from some $k$ on,
  for \eqref{eq:delta}, $\gamma:=k^{\lambda}$, and $N:=n_k$.

  According to \eqref{eq:main-1} and \eqref{eq:main-2},
  we will have, from some $k$ on,
  \begin{align*}
    S^k_{n_k}&\le\psi(n_k),\\
    -S^k_n&\le\psi(n_k-n_{k-1}) \text{ for all $n\le n_k-n_{k-1}$}.
  \end{align*}
  For any sufficiently large $N$,
  these inequalities imply the inequality in
  \begin{multline*}
    \sum_{n=-N}^{-1}
    (\xi_n-1)
    =
    \sum_{n=1}^{n_k}
    \eta_n^k
    -
    \sum_{n=1}^{n_k-N}
    \eta_n^k
    =
    S^k_{n_k} - S^k_{n_k-N}
    \le
    \psi(n_k)
    +
    \psi(n_k-n_{k-1})\\
    \sim
    \sqrt{2c \lambda n_k\ln\ln n_k}
    +
    \sqrt{2c \lambda(n_k-n_{k-1})\ln\ln(n_k-n_{k-1})},
  \end{multline*}
  where $k$ is the value satisfying $N\in[n_{k-1},n_k)$.
  Since we can take $\lambda$ arbitrarily close to 1,
  this completes the proof of \eqref{eq:LIL-upper}.

  To prove \eqref{eq:LIL-exact} for some bounded e-flow, suppose
  \[
    \xi_n
    =
    \begin{cases}
      C & \text{with probability $1/C$}\\
      0 & \text{with probability $1-1/C$}
    \end{cases}
  \]
  independently for $n=-1,-2,\dots$.
  Then $\var(\xi_n)=C-1$,
  and applying the standard law of the iterated logarithm
  gives~\eqref{eq:LIL-exact}.
\end{proof}


%% file: suppl_2.txt

In the proof of Proposition~\ref{prop:standard-LIL} we should use
the stopping times $\tau_k:=\min\{n:A_n\ge\lambda^k\}$
instead of the constant stopping times $n_k$.